\documentclass[sigconf]{acmart}

\AtBeginDocument{%
  }

\acmSubmissionID{3261}

\usepackage{orcidlink}
\usepackage{adjustbox}
\usepackage{array}
\usepackage{tabularx}
\usepackage{multirow}
\usepackage{caption}
\usepackage{float}
\usepackage{enumitem}
\setlist{nosep}
\usepackage{dblfloatfix}   
\usepackage{afterpage}     

\newcolumntype{M}[1]{>{\centering\arraybackslash}m{#1}}

\begin{document}

\title{SceneCraft: Interactive System for Image Editing via Scene Graph}

\author{Duc-Manh Phan}
\authornote{These authors contributed equally to this work.}
\email{22120205@student.hcmus.edu.vn}
\orcid{0009-0006-6648-3844}
\affiliation{%
  \institution{University of Science}
  \city{Ho Chi Minh}
  \country{Vietnam}}
\affiliation{%
  \institution{Vietnam National University}
  \city{Ho Chi Minh}
  \country{Vietnam}}

\author{Ngoc-Dai Tran}
\authornotemark[1]
\orcid{0009-0001-1541-9061}
\email{22120045@student.hcmus.edu.vn}
\affiliation{%
  \institution{University of Science}
  \city{Ho Chi Minh}
  \country{Vietnam}}
\affiliation{%
  \institution{Vietnam National University}
  \city{Ho Chi Minh}
  \country{Vietnam}}

\author{Duy-Khang Do}
\orcid{0009-0003-7430-1586}
\email{22120454@student.hcmus.edu.vn}
\affiliation{%
  \institution{University of Science}
  \city{Ho Chi Minh}
  \country{Vietnam}}
\affiliation{%
  \institution{Vietnam National University}
  \city{Ho Chi Minh}
  \country{Vietnam}}

\author{Tam V. Nguyen}
\orcid{0000-0003-0236-7992}
\email{tamnguyen@ydayton.edu}
\affiliation{%
  \institution{University of Dayton}
  \city{Dayton}
  \state{Ohio}
  \country{USA}}

\author{Minh-Triet Tran}
\orcid{0000-0003-3046-3041}
\email{tmtriet@fit.hcmus.edu.vn}
\affiliation{%
  \institution{University of Science}
  \city{Ho Chi Minh}
  \country{Vietnam}}
\affiliation{%
  \institution{Vietnam National University}
  \city{Ho Chi Minh}
  \country{Vietnam}}

\author{Trung-Nghia Le}
\authornote{Corresponding author.}
\orcid{0000-0002-7363-2610}
\email{ltnghia@fit.hcmus.edu.vn}
\affiliation{%
  \institution{University of Science}
  \city{Ho Chi Minh}
  \country{Vietnam}}
\affiliation{%
  \institution{Vietnam National University}
  \city{Ho Chi Minh}
  \country{Vietnam}}

\renewcommand{\shortauthors}{Manh, Tran, et al.}

\begin{abstract}
Recent advances in generative AI have enabled natural language-driven image editing, yet existing systems often fail in complex scenes with multiple interacting objects because they rely heavily on users crafting precise text prompts. To address the absence of structured control, we propose SceneCraft, a novel interactive framework that bridges user intent and model execution by representing images as editable scene graphs. Instead of guessing text prompts through trial and error, users interact directly with a visual graph to perform complex spatial and relational operations. These graph modifications are automatically translated into precise, context-aware editing prompts, effectively eliminating linguistic ambiguity. To ensure robust and diverse results, structured prompts are dispatched to multiple state-of-the-art generative models. Evaluations across diverse editing scenarios show that SceneCraft provides a more intuitive control mechanism, significantly reducing the cognitive burden of manual prompt engineering while generating outputs that users consistently rate as higher in quality and fidelity.
\end{abstract}


\begin{teaserfigure}
\centering
    \includegraphics[width=\textwidth]{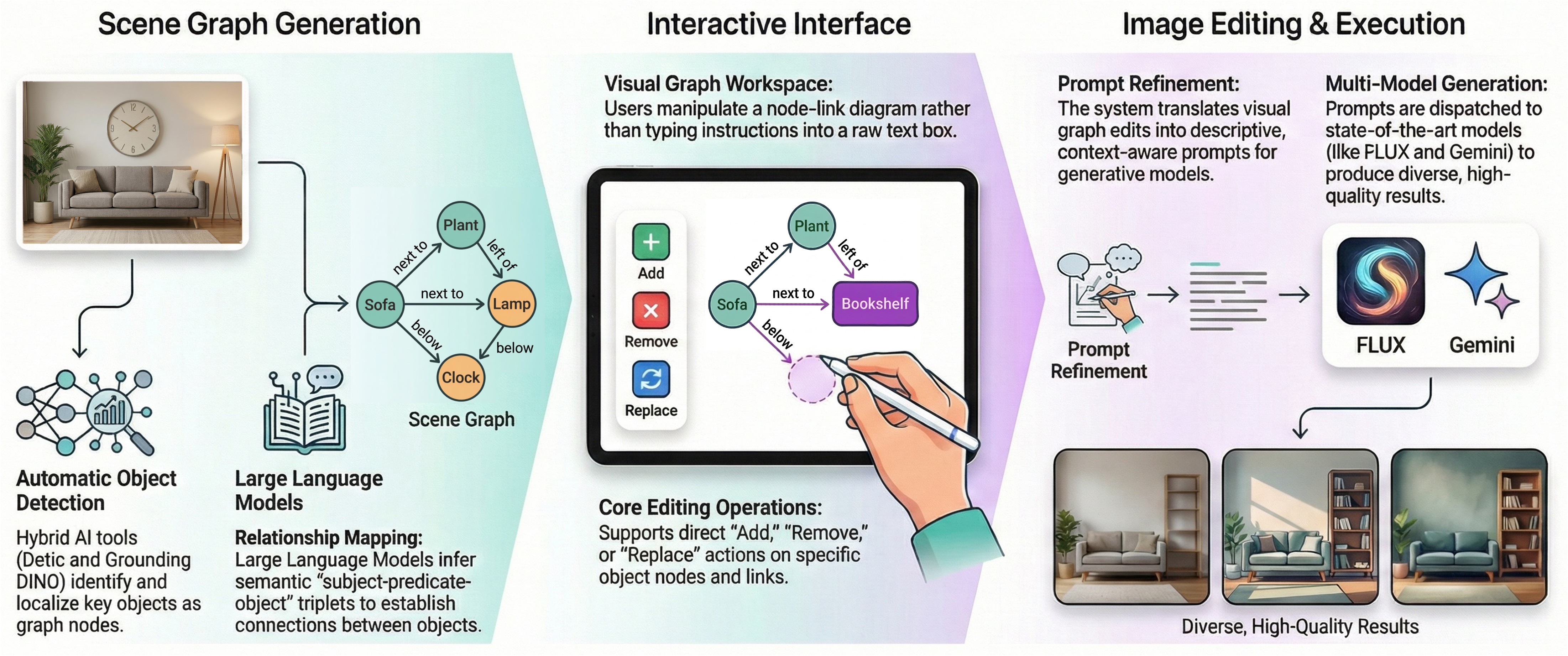}
    \caption{SceneCraft concept and interactive workflow. An input image is automatically parsed into a structured scene graph. Instead of relying on trial-and-error text prompting, users directly manipulate this visual graph by adding, removing, or replacing nodes. The system translates these graphical edits into structured prompts for generative models, reducing linguistic ambiguity and producing intent-aligned outputs.}
    \label{fig:full_pipeline}
\end{teaserfigure}


\maketitle

\section{Introduction}\label{sec1}

In recent years, generative AI has achieved significant progress in image synthesis and editing, offering powerful alternatives to traditional manual manipulation. Early work, such as InstructPix2Pix~\cite{brooks2023instructpix2pix}, demonstrated that diffusion models can follow natural language instructions for image editing, inspiring a growing body of instruction-driven approaches~\cite{fu2024mgie, kawar2023imagic,Vo2025CPAM}. More recently, advanced models such as FLUX.1 Kontext~\cite{labs2025flux1kontextflowmatching} improve consistency and robustness in image editing through flow matching and in-context learning, while multimodal systems including Qwen Image Editing~\cite{wu2025qwenimagetechnicalreport} and Gemini 2.5 Flash Image~\cite{google2025gemini25flashimage} leverage large multimodal language models to better interpret user instructions and guide the generation process. These developments have made image editing more accessible and flexible for a wide range of users.

\begin{figure}[t!]
\centering
\renewcommand{\arraystretch}{1}
\setlength{\tabcolsep}{2pt}
\begin{tabular}{>{\centering\arraybackslash}m{0.48\columnwidth}
                >{\centering\arraybackslash}m{0.48\columnwidth}}

\includegraphics[width=0.48\columnwidth, height=0.30\columnwidth]{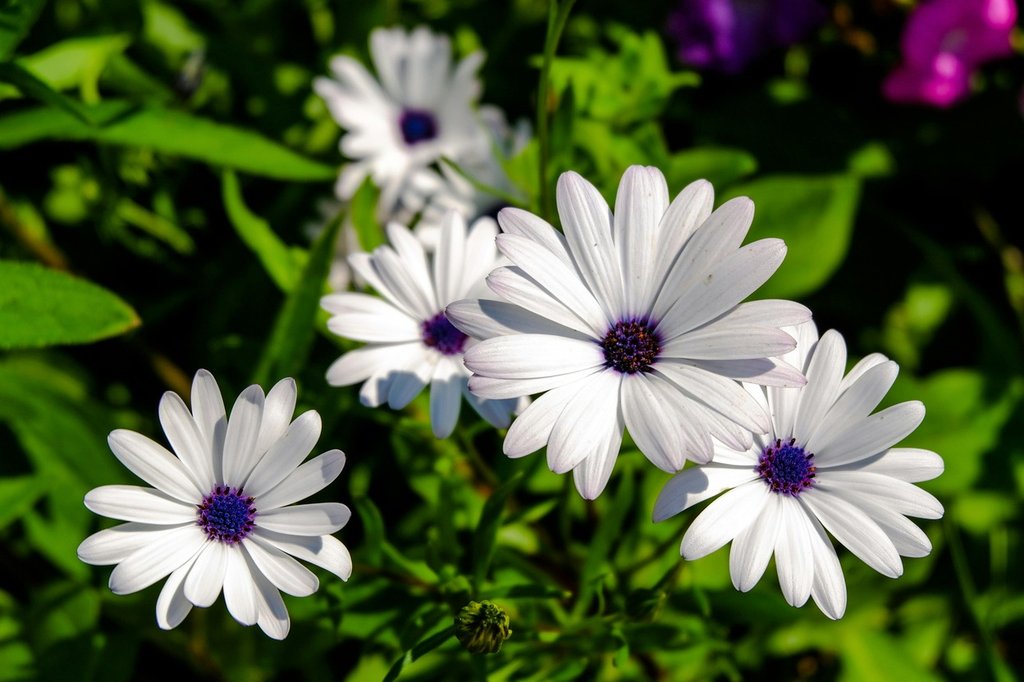}
& \includegraphics[width=0.48\columnwidth, height=0.30\columnwidth]{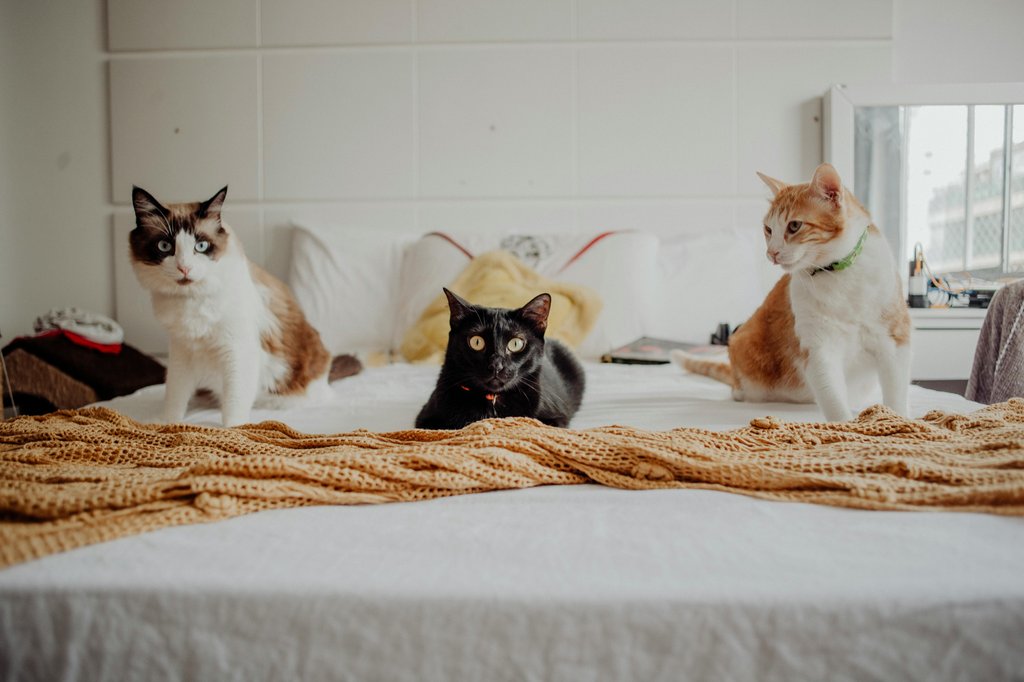} \\[6pt]

\includegraphics[width=0.48\columnwidth, height=0.30\columnwidth]{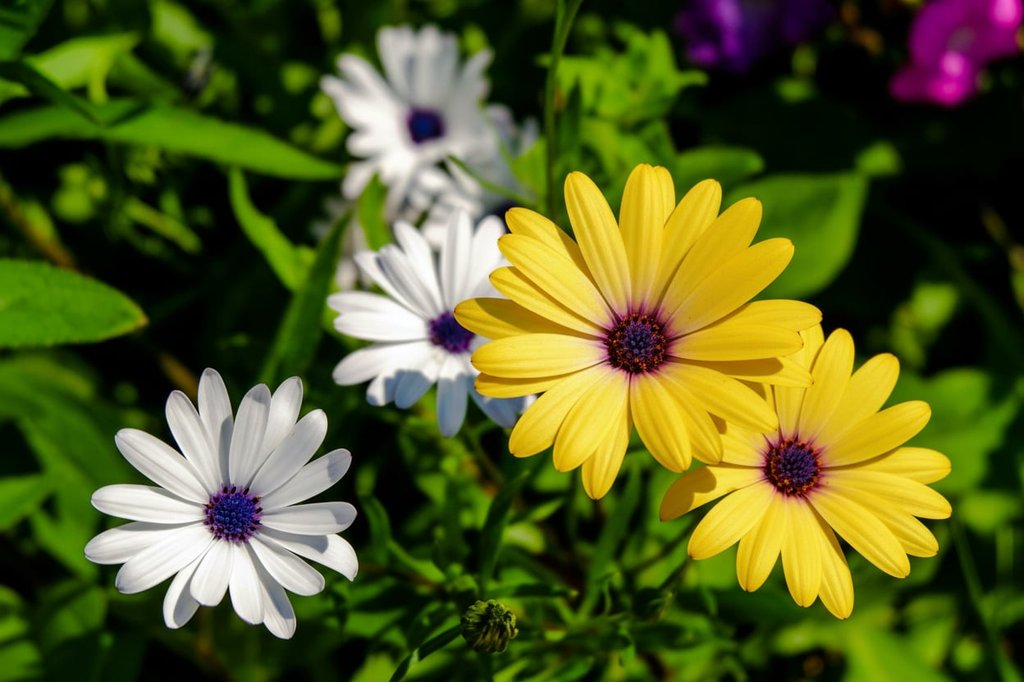}
& \includegraphics[width=0.48\columnwidth, height=0.30\columnwidth]{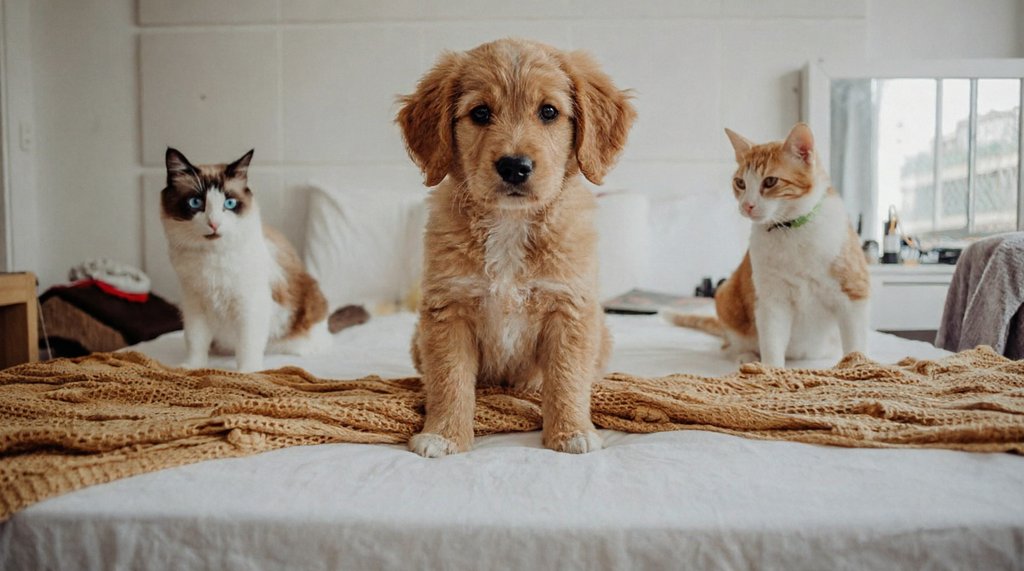} \\[4pt]

\parbox{0.47\columnwidth}{\centering\small\textit{turn the second flower from the right into a yellow one}}
& \parbox{0.47\columnwidth}{\centering\small\textit{replace the first cat from the left with a puppy}} \\

\end{tabular}
\caption{Limitations of existing text-driven image editing methods in multi-object scenes. The top row displays the original images, and the bottom row shows the corresponding edited images generated via standard text prompts.}
\label{tab:wrong_cases}
\vspace{-5mm}
\end{figure}

Despite this progress, existing image editing systems still face critical limitations (Fig.~\ref{tab:wrong_cases}). First, they struggle with complex scenes involving multiple interacting objects and relationships. Prior works such as SmartEdit~\cite{huang2024smartedit} attempt to improve instruction understanding using multimodal large language models, yet still lack explicit modeling of inter-object relationships. Meanwhile, SGEdit~\cite{zhang2024sgedit} introduces scene graph guidance to enable more structured editing, but relies on manually constructed graphs, limiting its practicality in real-world scenarios. These challenges are consistent with recent findings that current models remain limited in compositional reasoning and multi-object understanding. Second, current approaches rely heavily on user-crafted textual prompts. In practice, users often struggle to precisely articulate their editing intentions, especially when the desired changes involve complex spatial or relational constraints. MGIE~\cite{fu2024mgie} attempts to refine under-specified instructions using multimodal language models, and InsightEdit~\cite{xu2024insightedit} improves instruction following through better training strategies. However, these approaches still depend entirely on natural language input, without providing structured mechanisms to support users in expressing complex editing goals.

These shortcomings reveal a fundamental gap in the field, highlighting the absence of a structured and interactive representation capable of bridging user intent and model execution within complex, multi-object scenes. To address these interaction bottlenecks, we introduce \textbf{SceneCraft}, a novel interactive image editing framework that leverages editable scene graphs to guide the generation process (Fig.~\ref{fig:full_pipeline}). SceneCraft shifts the interaction paradigm away from raw text guessing and towards direct visual manipulation. Given an input image, our system automatically constructs a scene graph that captures objects and their relationships. Specifically, our system constructs this structured representation by employing hybrid AI detectors, combining Detic~\cite{zhou2022detecting} and  Grounding DINO~\cite{liu2023grounding}, to accurately localize key objects as graph nodes, and subsequently leverages a Large Language Model (LLM)~\cite{gemini25} to infer semantic subject-predicate-object relationships from the bounding box context to establish the connecting edges. Users can then interact directly with this structured representation, and their interactions are translated into editing prompts. This design reduces the burden of manual prompt engineering and enables more precise and interpretable control over complex edits.

Furthermore, standard text-to-image systems often produce homogeneous outputs that can limit creative exploration and ideation. To ensure robust execution and expand the visual output space, SceneCraft dispatches these structured prompts to multiple state-of-the-art generative models, including FLUX.1 Kontext~\cite{labs2025flux1kontextflowmatching}, Qwen Image Editing~\cite{wu2025qwenimagetechnicalreport}, and Gemini 2.5 Flash Image~\cite{google2025gemini25flashimage}. Because these models exhibit complementary strengths, our framework aggregates diverse, high-quality editing results. This mitigates single-model failure modes and allows users to select the outcome that best aligns with their vision.

We evaluate SceneCraft through user studies across diverse editing scenarios. Experimental results show that our framework provides more intuitive interaction and consistently improves user satisfaction. In summary, our contributions are threefold:
\begin{itemize}
\item We propose an interactive image editing framework based on scene graphs, reducing the need for complex prompt engineering.
\item We enhance scene graph construction by enriching object-level contextual information, improving relationship understanding in complex scenes.
\item We integrate multiple state-of-the-art generative models to increase robustness and provide diverse, high-quality editing results.
\end{itemize}

\section{Related Work}

\subsection{LLM-based Editing}
The integration of large language models (LLMs) into image editing has progressed from early instruction-driven pipelines to recent multimodal controllers. Early approaches such as InstructPix2Pix~\cite{brooks2023instructpix2pix} enabled free-form instruction-driven editing by retraining diffusion models on synthetic before–after pairs, but did not yet leverage external LLM reasoning. A key milestone was Visual ChatGPT~\cite{wu2023visualchatgpt}, which chained a conversational LLM with vision backbones (e.g., BLIP~\cite{li2022blip}, Stable Diffusion~\cite{rombach2022high}, ControlNet~\cite{zhang2023adding}) to decompose user requests into multi-step edits.

Subsequent work expanded LLMs beyond prompt following toward multimodal reasoning and spatial planning. Idea2Img~\cite{yang2023idea2img} used GPT-4V to iteratively refine prompts and select improved generations. Layout-centric methods (LMD~\cite{lian2023llm}, LayoutGPT~\cite{feng2024layoutgpt}) elicited layouts from LLMs to guide synthesis, while Attention Refocusing~\cite{phung2024attentionrefocusing}, RPG~\cite{yang2024mastering}, and SLD~\cite{wu2023self} employed LLM planning or self-correction loops to improve multi-object alignment and reduce prompt–image mismatch. Creative-VLA~\cite{peng2024creativevla} added chain-of-thought reasoning for translating abstract instructions into concrete edit operations.

Recent systems generalized LLM-guided editing to open domains and complex inputs. InstructAny2Pix~\cite{li2025instructany2pix} supports multimodal prompts, simultaneous multi-object edits, and external style references. SGEdit~\cite{zhang2024sgedit} highlights LLMs as both a scene parser and an editing controller, combining structured scene graphs with attention-modulated diffusion for precise object-level edits in complex scenes.

Concurrently, several {recent foundation editors} illustrate a trend toward integrating strong multimodal reasoning with generative backbones to improve robustness and usability in real-world editing. Among them, Qwen Image Editing~\cite{wu2025qwenimagetechnicalreport} and Gemini 2.5 Flash Image are {LLM-based multimodal editors}, where large language models directly parse and control editing. In contrast, FLUX~1 Kontext~\cite{labs2025flux1kontextflowmatching} remains a {flow-based} generator (Rectified Flow), but augments it with LLM-style reasoning to improve contextual understanding.

\subsection{User Interfaces for Generative AI}

A critical barrier to effective interaction is the unpredictability of model behavior and the lack of explainability in text-to-image pipelines. Crafting desired prompts presents difficulties, especially for beginner users who are unfamiliar with specialized or esoteric ``magic keywords.''

In response, various interactive UIs for generative AI have been developed. PromptMap~\cite{adamkiewicz2025promptmap} enables users to explore a vast synthetic collection of examples through a semantic map to find inspiration without tedious prompt engineering. POET~\cite{han2025poet} automatically detects and expands homogeneous dimensions in text-to-image models to diversify output spaces. GenAssist~\cite{huh2023genassist} supports blind and low-vision creators by utilizing LLMs to verify prompt alignment and extract visual styles. Similarly, Spiritus lowers the barrier for 2D character creation by combining sketch guidance with semantic-driven layered generation. 

Recent work has further explored intelligent interfaces that assist users in prompt ideation and iterative refinement. Promptify~\cite{brade2023promptify} provides an interactive workspace that supports prompt exploration through visual navigation, automatic suggestions, and clustering of generated images, enabling users to refine outputs without blind trial‑and‑error. PromptCrafter~\cite{baek2023promptcrafter} introduces a mixed-initiative dialogue framework in which users construct prompts step‑by‑step through conversations with an LLM, helping to clarify intent and avoid the need to discover esoteric ``magic keywords'' manually. These systems collectively aim to reduce the cognitive burden of prompt engineering by providing visualization, automated assistance, and structured guidance.

SceneCraft builds on this tradition by shifting the user's focus from guessing text prompts to directly manipulating a visual logic structure of the scene.

\section{Formative Study and Design Goals}

To deeply understand the friction users experience during image editing and to inform the design of SceneCraft, we conducted a formative study consisting of semi-structured interviews and observational tasks. We specifically sought to contrast the workflows and pain points of users highly accustomed to AI prompting versus those who rely on traditional direct-manipulation tools.

\subsection{Participants and Procedure}

We recruited 10 participants, including graduate students, undergraduate students, faculty, and researchers with experience in computer vision, computer graphics, or the use of visual design tools (e.g., illustration software and presentation tools). Participants in this group were familiar with creating visual content and working with related technologies. Their ages ranged from 18 to 40. We divided them into two distinct cohorts of 5 participants to capture a spectrum of interaction paradigms:

\begin{itemize}
    \item {Experts:} Frequent users of text-to-image generative models (e.g., ChatGPT and Gemini). These users generate and edit images via natural language prompts on a weekly basis.

    \item {Non-Experts:} Users who rarely or never use generative AI for image editing. Instead, they rely on conventional direct-manipulation tools (e.g., Adobe Photoshop, mobile photo editing apps) for their creative workflows.
\end{itemize}

We conducted a thematic analysis of the collected data, including interaction sessions, and follow-up interviews via open-ended questions. The 15-min study session is about participants' current image-editing practices, followed by a guided editing task where they were asked to perform specific modifications (e.g., adding, removing, and replacing objects) on complex, multi-object images. All participants were asked to use commercial LLM interfaces, such as Gemini or ChatGPT, for editing images.

\subsection{Key Insights}

Our thematic analysis of the interview transcripts and task observations revealed several critical bottlenecks in current image-editing workflows.

\textit{Insight 1: The ``black box'' of spatial prompting (expert focus).} Even for experts highly proficient in crafting prompts for ChatGPT or Gemini, precisely editing spatial relationships proved to be a severe bottleneck. Experts expressed frustration that these commercial tools treat complex spatial instructions as a ``black box''. When an expert prompted an LLM to “replace the cup on the left with a vase, but keep the plate behind it,” the model frequently failed to parse the spatial constraints accurately, leading to hallucinations or incorrect target modifications. As one expert noted, ``If I pin down something really specific or narrow, AI seems to break down.'' Because models often lack explicit structural transparency, experts are forced into a tedious trial-and-error loop, continuously tweaking prepositions and adjectives in hopes that the model will eventually align with their spatial intent.

\begin{figure}[t]
    \centering
    \includegraphics[width=\linewidth]{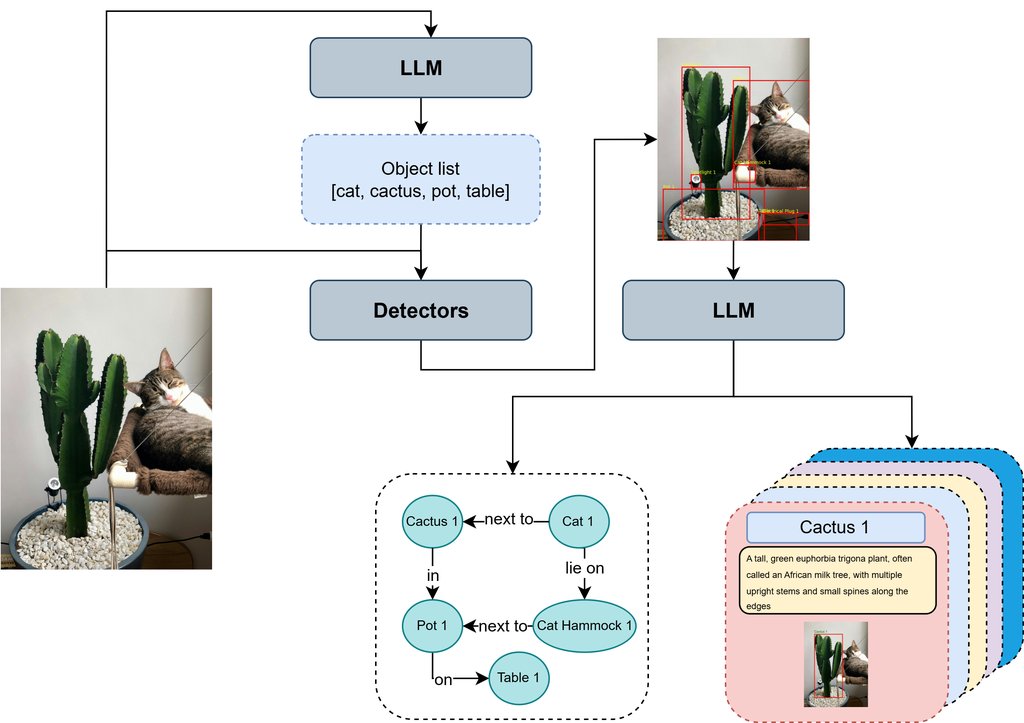}
    \caption{Scene graph generation pipeline. Given an input image (left), semantic labels for the main objects are first extracted and passed to detectors (e.g., Detic~\cite{zhou2022detecting}, Grounding DINO~\cite{liu2023grounding}) to provide candidate regions. The LLM then generates objects and infers pairwise relationships based on the bounding box context, producing a structured scene graph and detailed object descriptions (right).}
    \label{fig:sg_gen}
\end{figure}

\begin{figure*}[t!]
\centering
\renewcommand{\arraystretch}{1}
\setlength{\tabcolsep}{2pt}
\begin{tabularx}{\textwidth}{
  >{\centering\arraybackslash}m{0.03\textwidth}  
  >{\centering\arraybackslash}m{0.26\textwidth}  
  >{\centering\arraybackslash}m{0.17\textwidth}  
  >{\centering\arraybackslash}m{0.26\textwidth}  
  >{\centering\arraybackslash}m{0.16\textwidth}  
}

 &
\includegraphics[width=\linewidth]{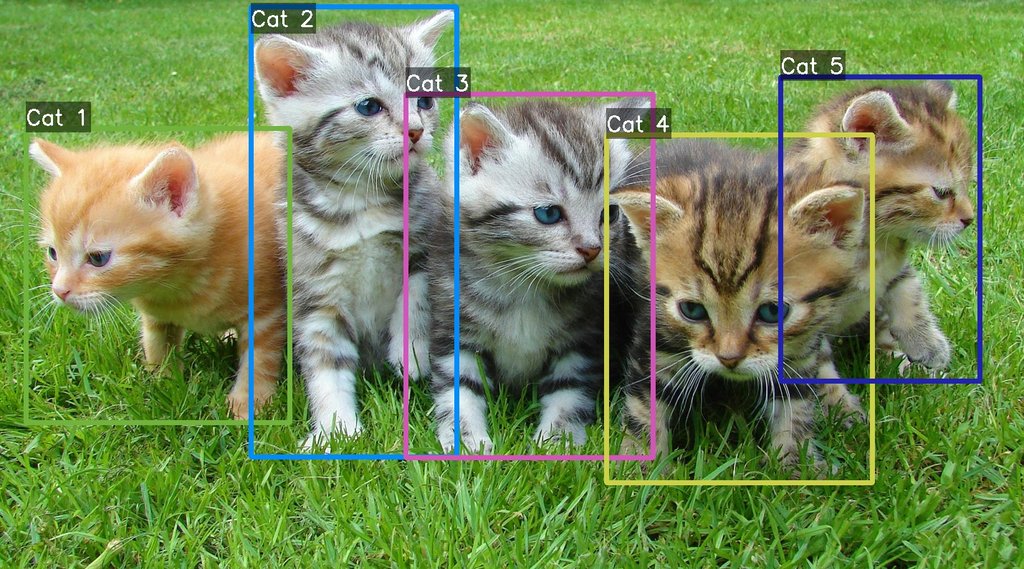} 
& \includegraphics[width=\linewidth]{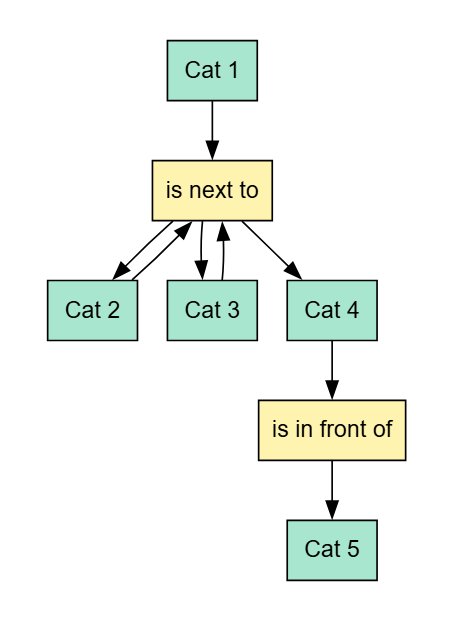} 
& \includegraphics[width=\linewidth]{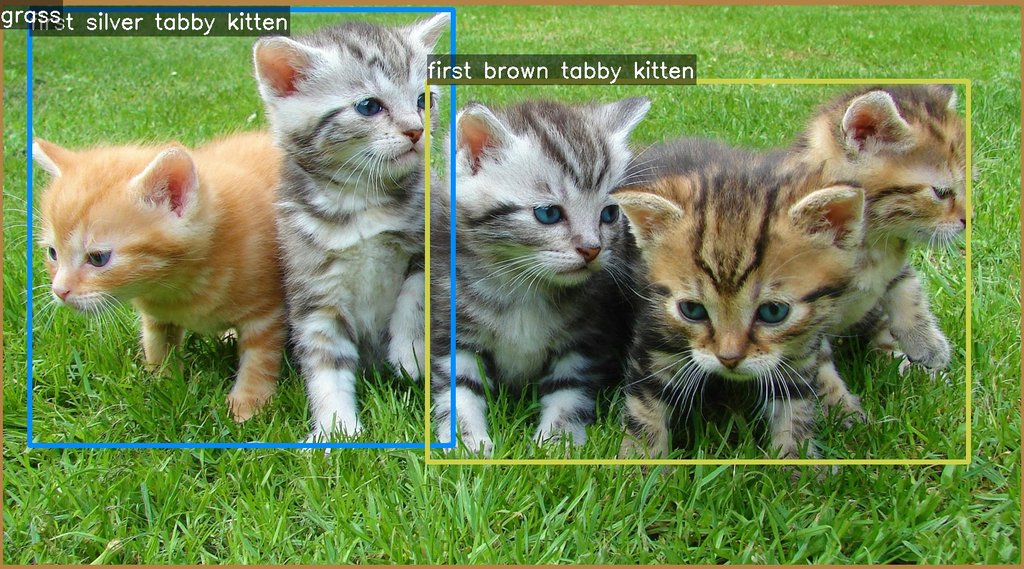} 
& \includegraphics[width=\linewidth]{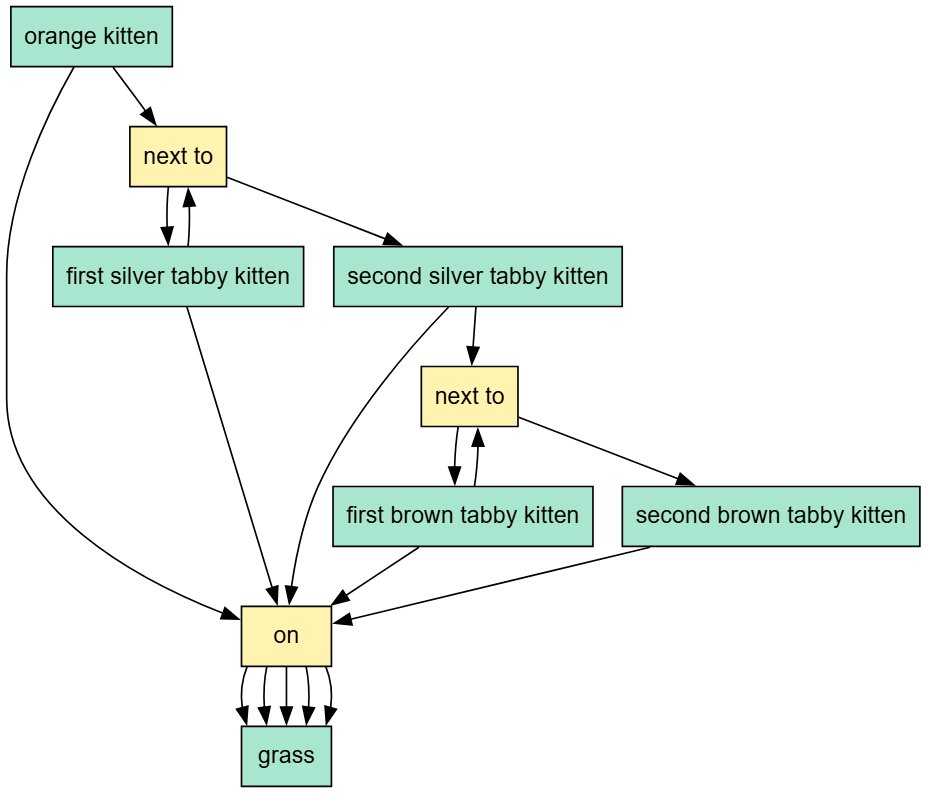} \\



 &
\includegraphics[width=\linewidth]{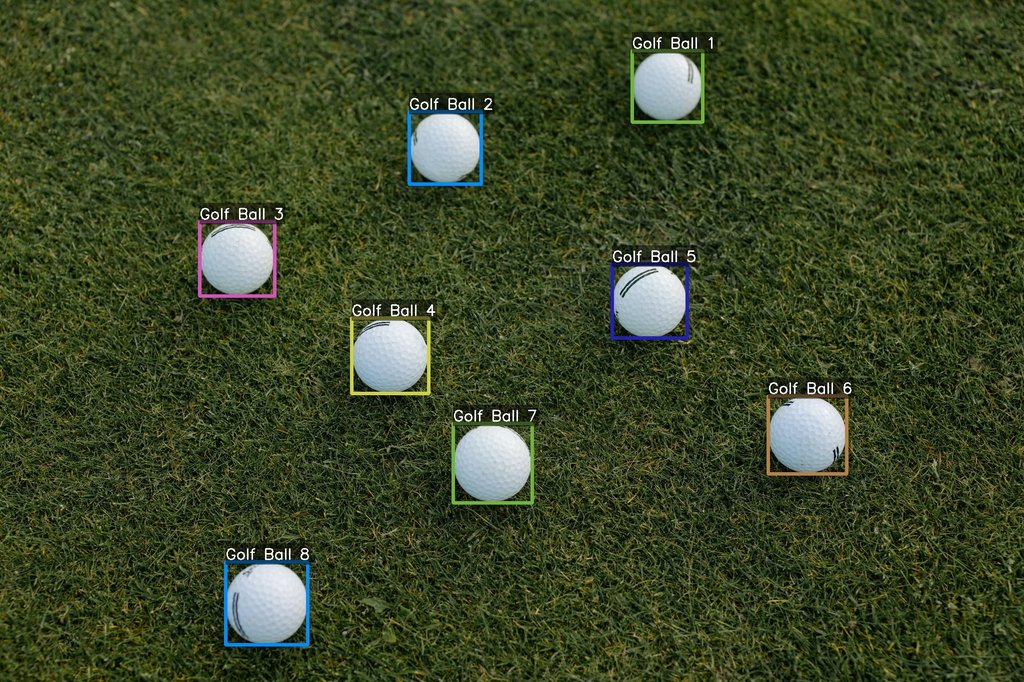} 
& \includegraphics[width=\linewidth]{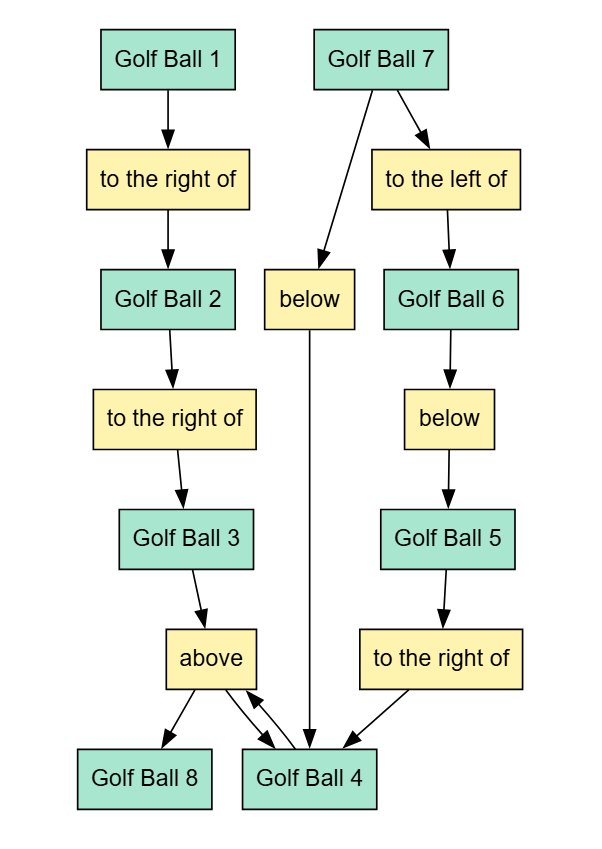} 
& \includegraphics[width=\linewidth]{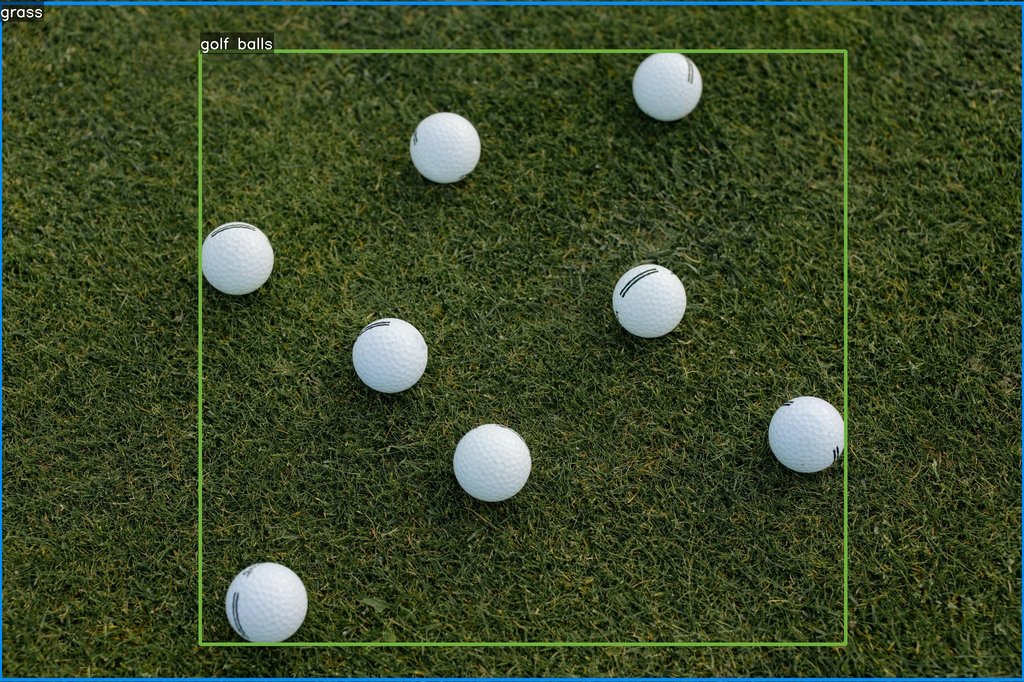} 
& \includegraphics[width=\linewidth]{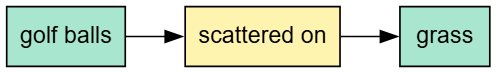} \\

&
\multicolumn{2}{c}{a) {SceneCraft}} 
& \multicolumn{2}{c}{b) {SG-Edit}} \\
\end{tabularx}
\caption{Visual comparison of scene graph generated from our SceneCraft and SG-Edit~\cite{zhang2024sgedit}. SG-Edit often fails to detect all objects properly because object names are fed directly as prompts to SAM, which can lead to incomplete detection or merged bounding boxes (masks), whereas SceneCraft's hybrid detection approach ensures accurate localization.}
\label{fig:scenecraft-sgedit}
\end{figure*}

\textit{Insight 2: The disconnect between linguistic guesswork and direct manipulation (non-expert focus).} Non-experts experienced a steep learning curve when attempting to translate their visual goals into text prompts. Because they were accustomed to the direct manipulation afforded by conventional tools like Photoshop (where objects are easily selected and moved on independent layers), they found raw text-prompting highly unnatural for structural edits. Traditional 2D workflows rely on manual layering and direct selection, whereas generative AI models typically output ``non-layered and cannot be directly used in professional creative workflows.'' Non-experts desired the automation power of AI but demanded the explicit, component-level control they were used to in conventional UI software.

\begin{figure*}[t!]
    \centering
    \includegraphics[width=\textwidth]{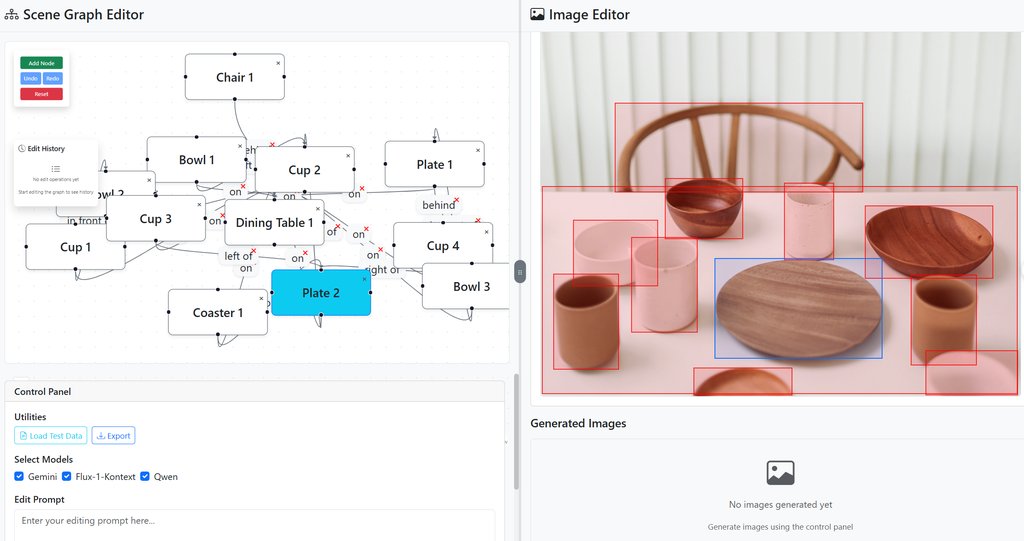}
    \caption{SceneCraft interactive editing interface. Users can select and modify nodes in the Scene Graph Editor (left), and edits are automatically translated into precise prompts for the image editing models, updating the Image Editor (right).}
    \Description{Screenshot of the SceneCraft interactive web interface showing scene graph editing and image editing controls.}
    \label{fig:user_interface}
\end{figure*}

\textit{Insight 3: The burden of iteration and ``prompt fragility.''} Both groups suffered from what we term prompt fragility. When experts attempted to fix a single object in a generated scene by modifying a few words in their text prompt, the generative model would often drastically alter the entire image background or the style of unrelated objects. Because raw text does not inherently lock down the ``scene logic,'' users felt a lack of control over the consistency of their edits. Participants desired a way to structurally ``freeze'' the relationships of certain objects while exclusively modifying others without relying on complex, esoteric ``magic keywords.''

\subsection{Design Goals}

Based on these formative insights, we derived three core Design Goals (DGs) to guide the development of the SceneCraft interface:

\begin{itemize}
    \item \textbf{DG1:} Abstract complex prompt engineering. Users should not need to master esoteric prompt keywords.
    \item \textbf{DG2:} Provide intuitive spatial and relational control. The system must allow direct manipulation of multi-object interactions.
    \item \textbf{DG3:} Support  alternative explorations. The system should offer diverse outputs from multiple models to mitigate single-model failure modes
\end{itemize}

\section{Proposed System}

\subsection{System Overview}

As illustrated in Fig.~\ref{fig:full_pipeline}, our system takes an image as input, automatically extracts objects and their spatial/semantic relationships to produce a structured scene graph, which can then be interactively edited and used to guide image editing. SceneCraft consists of three main components: (1) scene graph generation, where objects and their relationships are extracted; (2) an interactive interface that allows user modifications; and (3) an image editing module that executes edits based on refined, LLM-generated prompts.

\subsection{Scene Graph Generation}

To relieve users from the burden of manually defining structural logic, our system transforms the input image into a structured scene graph automatically. We construct a scene graph $G(O,R)$, where $O = \{o_1, o_2, \ldots, o_n\}$ is the set of objects and $R = \{r_1, r_2, \ldots, r_m\}$ is the set of relationships. 
Each relation is represented as a triplet $(o_i, r_k, o_j)$, corresponding to a subject–predicate–object structure. The generation process is decomposed into two steps as illustrated in Fig.~\ref{fig:sg_gen}.

\subsubsection{Main Object Detection}

We introduce a hybrid detection–LLM approach to parse the scene accurately. First, Detic~\cite{zhou2022detecting} generates candidate bounding boxes for all possible objects. Simultaneously, an LLM~\cite{gemini25} is prompted to identify a structured list of the main objects. Each object is then precisely localized by Grounding DINO~\cite{liu2023grounding} based on text queries. This hybrid approach allows Detic and Grounding DINO to complement each other: Detic provides broad coverage and Grounding DINO delivers precise localization resolves over-segmentation issues, overcoming incomplete detection or merged masks of previous methods~\cite{zhang2024sgedit}. The bounding boxes are merged via IoU-based matching, and the LLM refines the results by assigning semantic labels and unique IDs (e.g., kitten 1, ball 1), forming the graph nodes.

\subsubsection{Relationship Generation}

Given the object set $O$ with their bounding boxes and the original image, we prompt the LLM~\cite{gemini25} to infer semantic relationships. This step establishes the edges of the scene graph, resulting in a structured representation of the scene.

While recent approaches like SG-Edit~\cite{zhang2024sgedit} have introduced scene graph guidance to enable structured image editing, they often rely on manually constructed graphs or struggle with precise object localization in complex multi-object scenes. Specifically, SG-Edit directly feeds object names as textual prompts into segmentation models like Segment Anything (SAM)~\cite{kirillov2023segment}, a method that frequently results in incomplete detections or erroneously merged bounding boxes and masks when multiple interacting objects are present. In contrast, SceneCraft overcomes these limitations through a robust hybrid detection approach that combines Detic for broad candidate coverage and Grounding DINO for precise, text-queried localization. As demonstrated in Fig.~\ref{fig:scenecraft-sgedit}, SceneCraft's pipeline ensures that all individual objects are accurately detected and distinctly separated, successfully avoiding the merged mask failures prevalent in SG-Edit and providing a significantly more reliable structural foundation for user-driven relational edits.

\begin{figure*}[t!]
\centering
\renewcommand{\arraystretch}{1.2}
\setlength{\tabcolsep}{3pt}
\begin{tabularx}{\textwidth}{
  >{\centering\arraybackslash}m{0.20\textwidth}
  >{\raggedright\arraybackslash}m{0.10\textwidth}
  >{\centering\arraybackslash}m{0.20\textwidth}
  >{\raggedright\arraybackslash}m{0.10\textwidth}
  >{\centering\arraybackslash}m{0.20\textwidth}
  >{\raggedright\arraybackslash}m{0.10\textwidth}
}

\includegraphics[width=\linewidth]{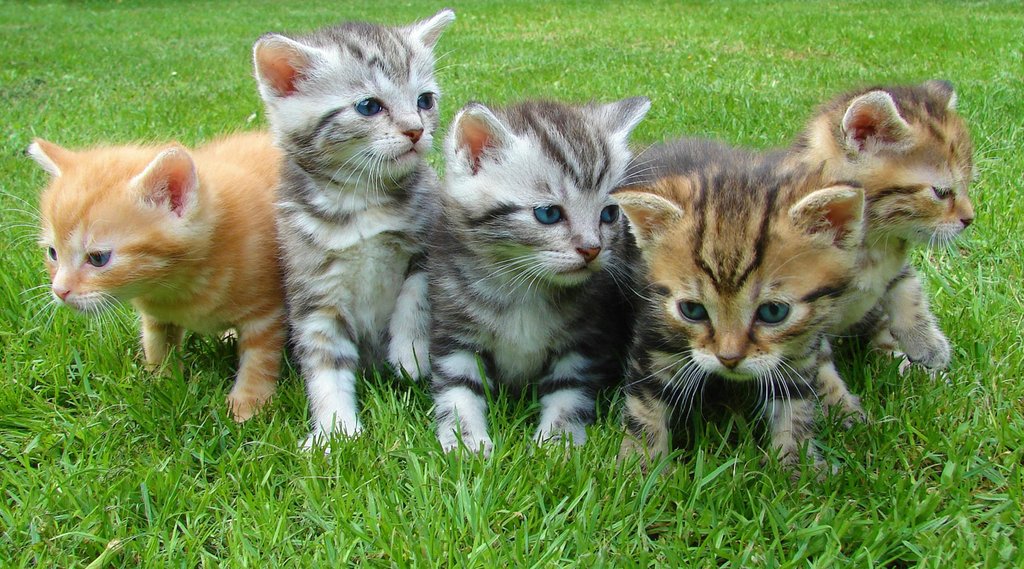} &
{\tiny \textbullet~Remove the smallest kitten on the left.\newline\textbullet~Add a red ball in front of the kittens.} &
\includegraphics[width=\linewidth]{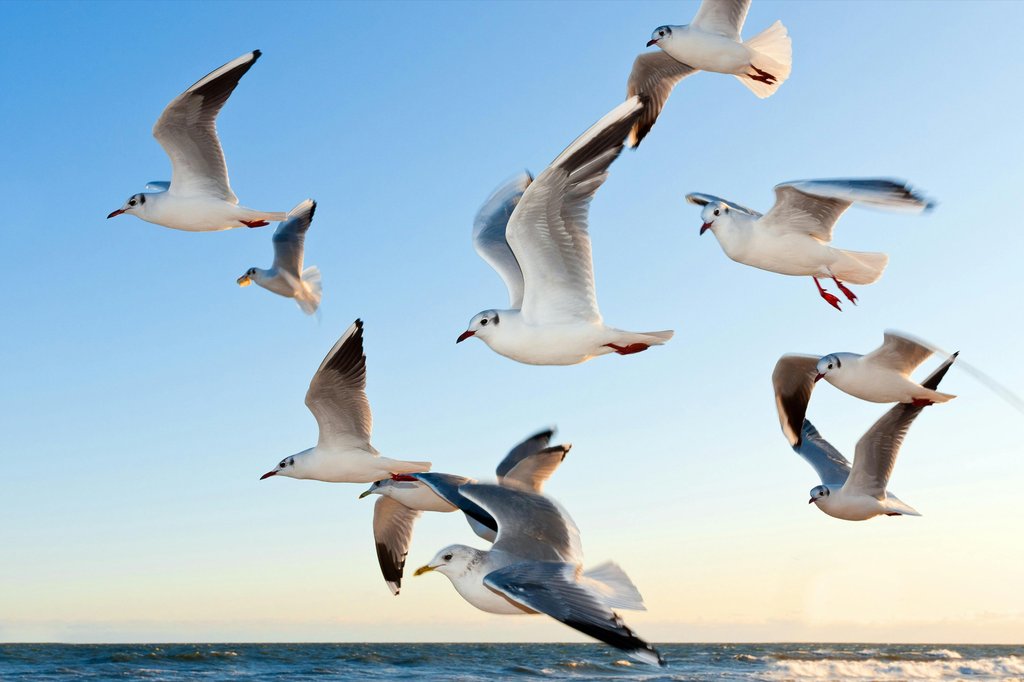} &
{\tiny \textbullet~Remove the two seagulls at the top.\newline\textbullet~Replace the highest-flying seagull with a hawk.} &
\includegraphics[width=\linewidth]{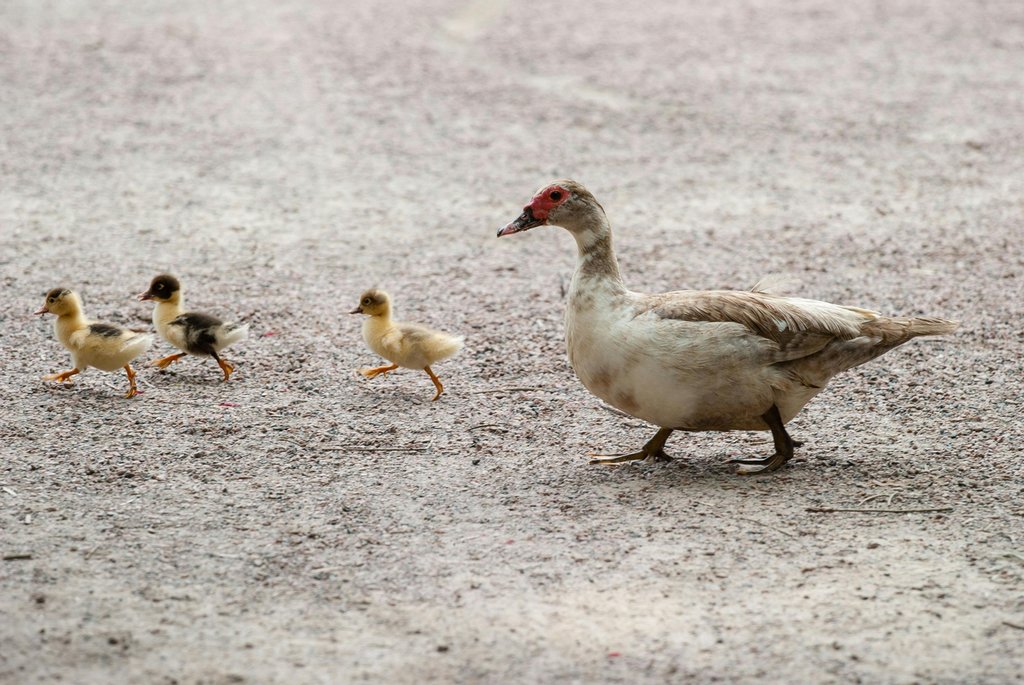} &
{\tiny \textbullet~Add a small pond behind the ducks.\newline\textbullet~Replace the mother duck with a white hen.} \\[4pt]

\includegraphics[width=\linewidth]{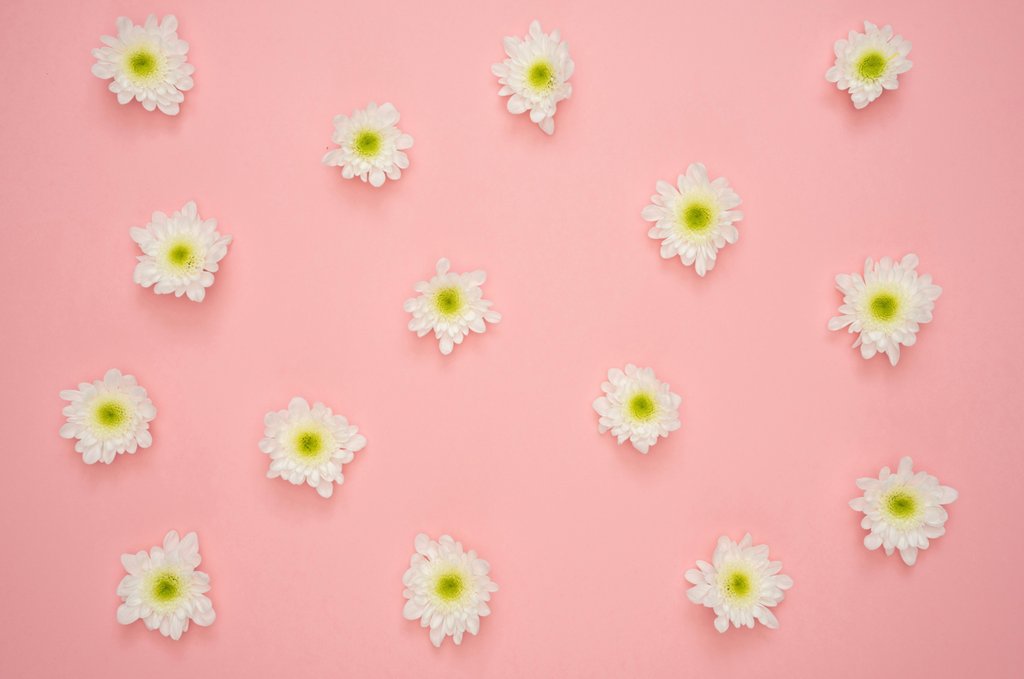} &
{\tiny \textbullet~Remove the three flowers in the top-left corner.\newline\textbullet~Add a green leaf in the center.} &
\includegraphics[width=\linewidth]{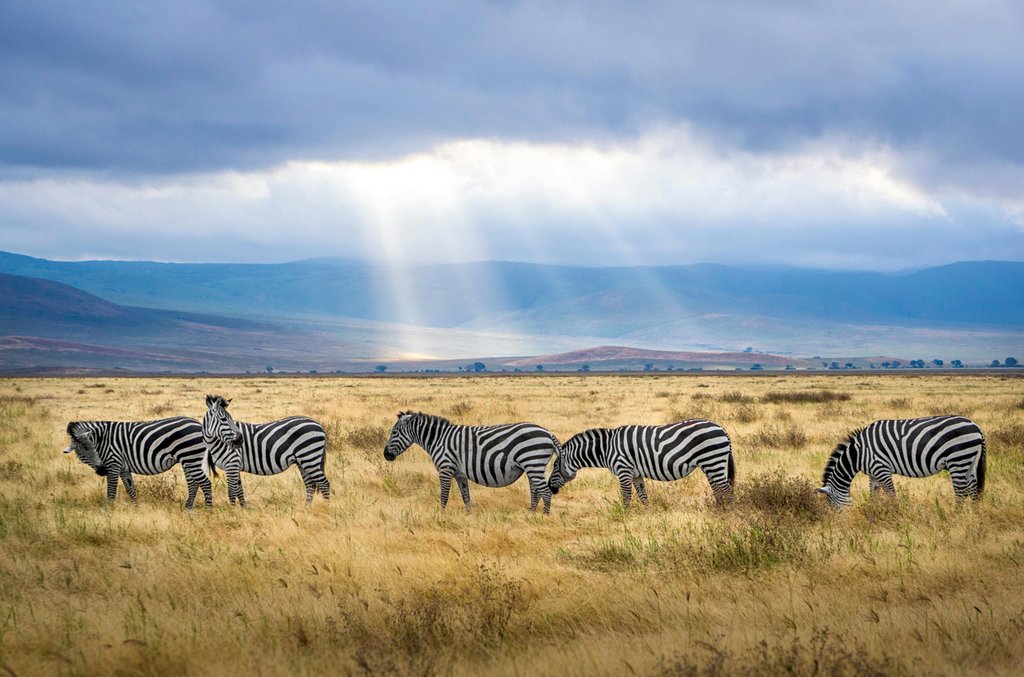} &
{\tiny \textbullet~Remove the third zebra from the left.\newline\textbullet~Replace the rightmost zebra with a horse.} &
\includegraphics[width=\linewidth]{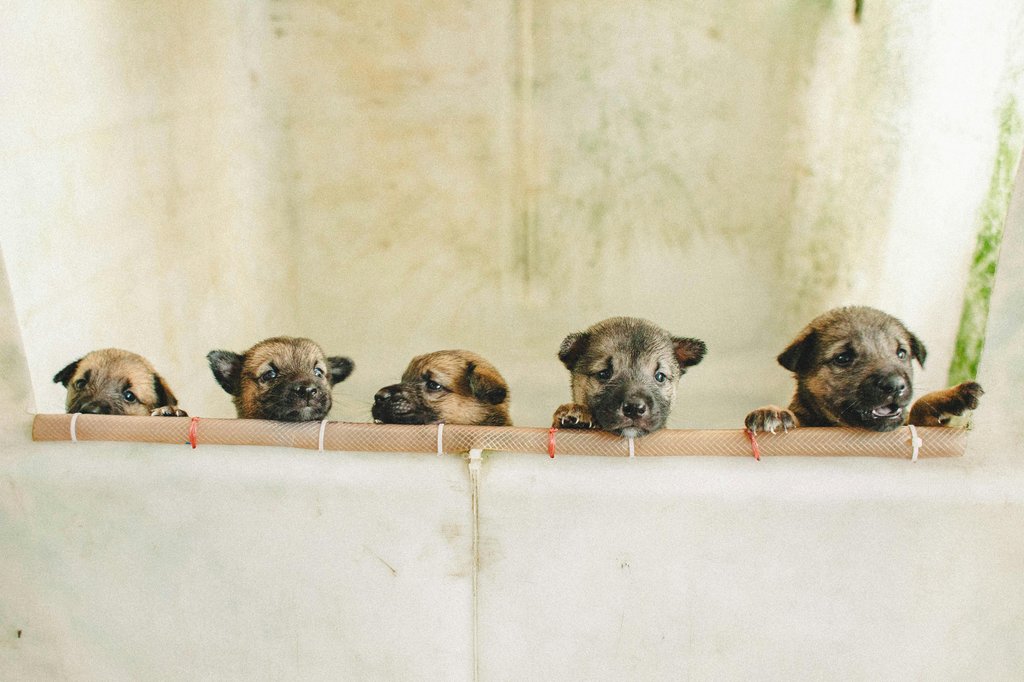} &
{\tiny \textbullet~Remove the second puppy from the left.\newline\textbullet~Add a ball in front of the leftmost puppy.\newline\textbullet~Replace the rightmost puppy with a kitten.} \\[4pt]

\includegraphics[width=\linewidth]{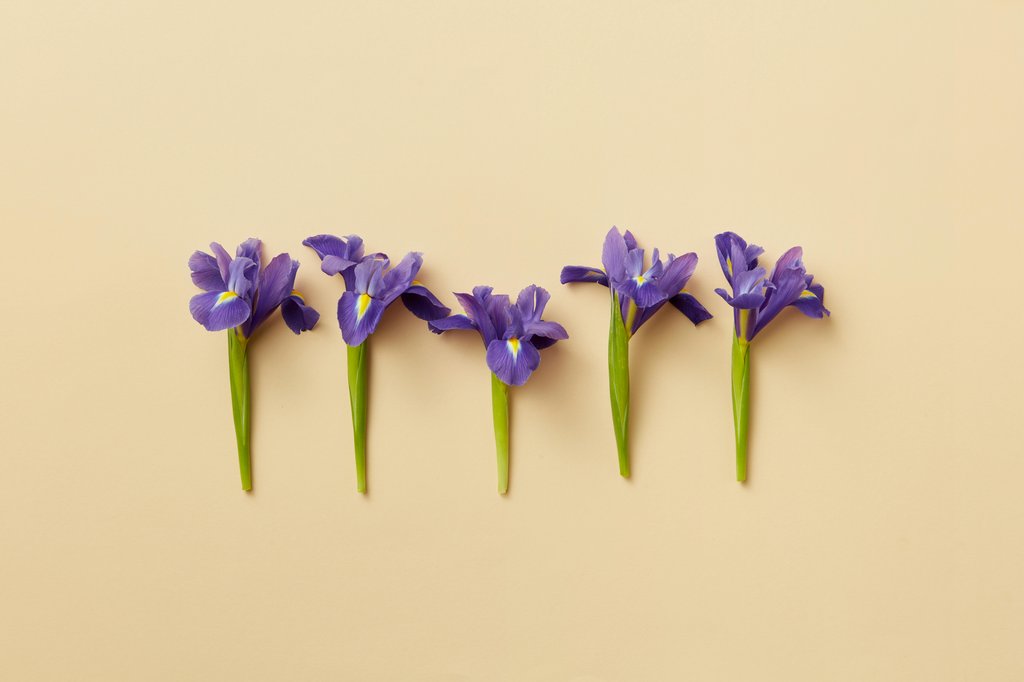} &
{\tiny \textbullet~Remove the two irises on the right.\newline\textbullet~Add a butterfly above the middle iris.\newline\textbullet~Replace the leftmost iris with a tulip.} &
\includegraphics[width=\linewidth]{images/orignial_3.jpg} &
{\tiny \textbullet~Remove the daisy on the far left.\newline\textbullet~Add a bee on the central daisy.\newline\textbullet~Replace the largest daisy with a sunflower.} &
\includegraphics[width=\linewidth]{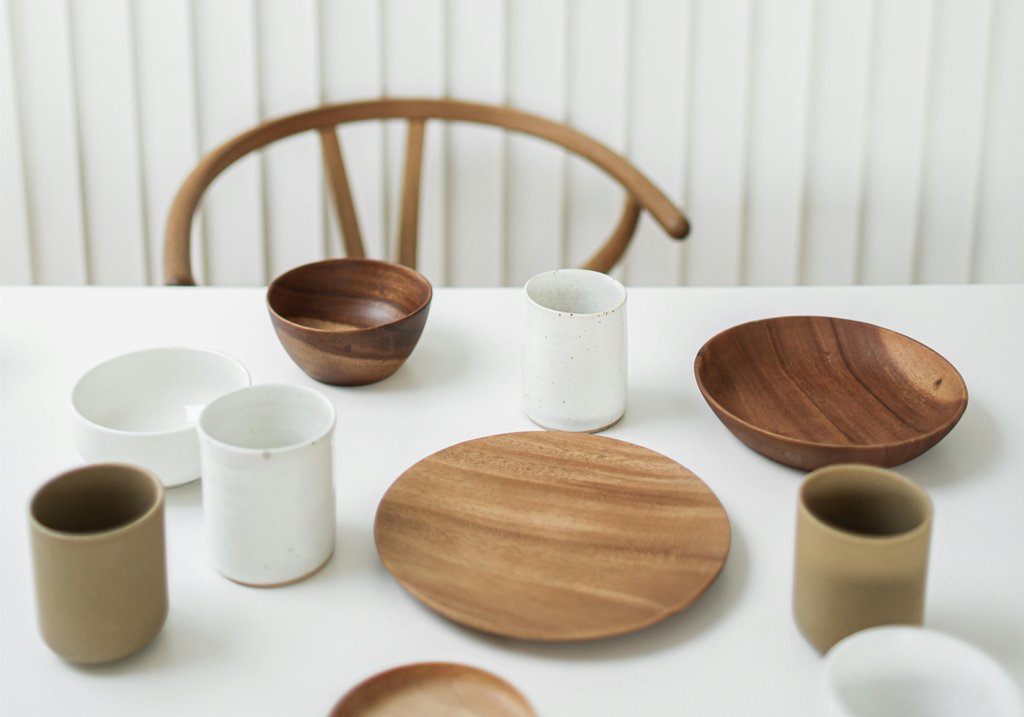} &
{\tiny \textbullet~Remove the small cup on the right side.\newline\textbullet~Add a spoon next to the round plate.\newline\textbullet~Turn the largest bowl to a ceramic vase.} \\

\end{tabularx}
\caption{Dataset samples and corresponding task descriptions. Our dataset contains multiple objects in both simple scenes (low object interaction) and complex scenes where objects overlap or maintain explicit spatial/semantic relations.}
\label{fig:dataset_samples}
\end{figure*}

\subsection{Interactive Workspace}

SceneCraft transitions users away from raw text boxes into a visual workspace featuring a Scene Graph Editor and an Image Editor (Fig.~\ref{fig:user_interface}).

\textit{Visual Integration:} When a user uploads an image (e.g., a dining table with multiple cups and plates), the Image Editor displays the original image overlaid with the generated bounding boxes. Simultaneously, the Scene Graph Editor visualizes a node-link diagram representing the semantic logic of the scene (e.g., Cup 2 → on → Dining Table 1).

\textit{Direct Graph Manipulation:} Instead of typing ambiguous spatial instructions (e.g., ``remove the cup next to the plate''), the user directly edits the graph representation. The system supports three core operations through direct manipulation:
\begin{itemize}
    \item Remove: Users select an object node and click ``delete''. The system enforces the preservation of the background context.

    \item Add: Users click the ``Add Node'' button, define a new object, and drag a relational link (an edge) to an existing anchor node in the graph to lock in its spatial location.
    
    \item Replace: Users double-click a node to substitute it with a new concept while explicitly maintaining the layout, relational context, and lighting of the original node.
\end{itemize}

\subsection{Prompt Translation \& Multi-Model Execution}

Once the user modifies the scene graph, the interface acts as a bridge to the generative backbones.

\textit{Context-Aware Prompt Refinement:} Every graphical interaction is translated into a raw editing instruction. For example, if a user deletes the node \textit{Cat 1}, the system generates the raw instruction \textit{``Delete Cat 1''}. This instruction, paired with the full structured scene graph, is passed to a LLM~\cite{gemini25}. The LLM refines this into a highly descriptive, context-aware prompt that specifies the target object, its precise location, and the surrounding relational context, effectively eliminating linguistic ambiguity.

\textit{Multi-Model Generation:} To maximize robustness and support diverse creative exploration (DG3), these structured prompts are dispatched to three state-of-the-art models: FLUX.1 Kontext~\cite{labs2025flux1kontextflowmatching}, Qwen Image Editing ~\cite{wu2025qwenimagetechnicalreport}, and Gemini 2.5 Flash Image~\cite{google2025gemini25flashimage}. Because these models have complementary strengths, this allows the user to review a gallery of diverse interpretations and select the highest quality result.

\section{User Study}

To further understand how users perceive SceneCraft compared to traditional raw prompting, we conducted a comprehensive within-subjects study evaluating both technical output fidelity and user experience

\subsection{Participants}

We recruited another group of 20 students aged 18–25 (12 male, 8 female). This comprises two subgroups, namely, 10 participants with prior experience in AI or design-related projects and 10 participants with no prior experience.
\subsection{Dataset}

We curated a dataset of 20 images collected from the Internet, each containing 2--6 foreground objects spanning animals, vehicles, and household items (Fig.~\ref{fig:dataset_samples}). We intentionally include both \emph{simple} scenes with minimal object interaction (e.g., kittens sitting on grass, iris flowers arranged in a row) and \emph{complex} scenes where multiple objects overlap or maintain explicit spatial/semantic relations (e.g., a dozen daisy flowers scattered across a pink background, zebras standing and grazing together on a savanna, wooden bowls and cups arranged together on a white surface).

\subsection{Benchmark Methods}

We compare {SceneCraft (ours)} against three state-of-the-art training-free editors:
{Qwen Image Editing}, {FLUX~1 Kontext}, and {Gemini~2.5 Flash Image}.
All experiments run on NVIDIA T4 GPUs.
For each edit, SceneCraft converts the scene-graph interaction into a structured prompt and queries the three backbone editors; the same instruction is also issued to each backbone as a raw user-written prompt for a fair comparison.

\subsection{Tasks}

We consider three standard editing operations:
\begin{itemize}
    \item {Remove}: delete a selected object while preserving background content.
    \item {Add}: insert a new object with plausible placement and realistic blending.
    \item {Replace}: substitute an object with another while maintaining layout, context, and lighting.
\end{itemize}

We note that an image can have different tasks with different requirements (Fig.~\ref{fig:dataset_samples}).

\subsection{Evaluation Metrics}

Since there is no ground truth for open-ended image editing, we evaluate outputs based on three criteria:
{Element Composition (EC)}, {Relationship Alignment (RA)}, and {Image Quality (IQ)}.

\paragraph{Element Composition (EC)}
Measures whether the edited image preserves the visual identity and counts of unaffected objects while correctly realizing the specified edit. Participants use a checklist derived from the scene graph (items to \emph{preserve}, \emph{add}, \emph{remove}) and mark each item \emph{correct}/\emph{incorrect}.


\paragraph{Relationship Alignment (RA)} Measures the fraction of scene-graph triples (subject–predicate–object) correctly satisfied in the edited output. Participants verify each triple, and RA is computed as the proportion of triples that are correctly satisfied.


\paragraph{Image Quality (IQ)}
Assesses overall realism (object fidelity, texture, lighting) and seamless background blending.


For pairwise comparison between methods $A$ and $B$, raters choose the preferred output (ties allowed).
The winning rate is computed separately for EC, RA, and IQ (ties excluded):
\[
\text{WinRate}(A\!:\!B) = \frac{\#\text{wins of }A}{\#\text{wins of }A + \#\text{wins of }B}\times 100\%,
\]

We also report 1--5 Likert mean opinion score (MOS) for each method, each metric independently.

\subsection{Apparatus and Procedure}

The pilot study was conducted both online and in-person to ensure a diverse participant pool. To familiarize participants with the evaluation workflow, online users received detailed instructions and a guided tutorial session, while in-person participants were provided with direct, real-time assistance as needed. All collected data was securely stored and subsequently analyzed across each metric and method combination.

During each trial, participants were presented with three primary components: (i) the original input image; (ii) the scene-graph edit, representing the user's intent as a structural delta on the scene graph (e.g., removing the node \texttt{zebra\#3}, adding the node \texttt{cloth} with the relation \texttt{under(eyeglasses, cloth)}, or replacing \texttt{cat\#1} with \texttt{dog}), rendered as a concise operator text alongside a small scene graph thumbnail; and (iii) the four edited outputs produced by {SceneCraft, Qwen Image Editing, FLUX~1 Kontext, Gemini~2.5 Flash Image}.

To keep the study duration manageable while comprehensively covering all core operations (remove, add, and replace), we randomly sampled 12 edit cases per participant from our curated 20-image dataset. Across our 20 participants, this within-subjects design yielded a total of $12 \times 20 \times 3 = 720$ pairwise preferences (3 baseline comparison pairs per case) and a corresponding set of Mean Opinion Score (MOS) ratings from the simultaneous 4-up evaluation task. To mitigate any potential learning or layout biases, both the presentation sequence of the images and the display order of the generative methods were strictly randomized across all participants and trials.


\subsection{Generated Content Evaluation}

\subsubsection{Quantitative Results}

For each edit, we evaluated outputs from SceneCraft and raw-prompt baselines (the same backbones driven directly by user-written prompts). Table~\ref{tab:mos} shows that SceneCraft achieved the highest MOS across all criteria. SceneCraft achieved a dominant 4.2 in EC, 4.1 in RA, and 4.4 in IQ, significantly outperforming the baseline models (which scored between 3.1 and 3.9), confirming that scene-graph guidance improves \emph{what} is edited (EC), \emph{how} objects relate after editing (RA), and the overall realism (IQ). In pairwise comparisons excluding ties (Table~\ref{tab:winrate}), SceneCraft achieved a 71.0\% to 77.3\% winning rate in EC, a 69.8\% to 75.1\% winning rate in RA, and a 68.5\% to 74.2\% winning rate in IQ. Scene-graph guidance drastically reduced semantic drift and improved background preservation compared to raw prompting.

\begin{table}[t!]
\centering
\caption{Evaluation results of objective image quality metrics, using Mean Opinion Scores (MOS; 1–5 scale). SceneCraft achieves the highest scores in Element Composition, Relationship Alignment, and overall Image Quality compared to raw-prompt baselines.}
\vspace{-2mm}
\label{tab:mos}
\begin{tabular}{lcccc}
\toprule
\textbf{Method} & \textbf{EC}\(\uparrow\) & \textbf{RA}\(\uparrow\) & \textbf{IQ}\(\uparrow\) \\
\midrule
Qwen Image Editing     & 3.5 & 3.4 & 3.9  \\
FLUX~1 Kontext         & 3.3 & 3.2 & 3.7  \\
Gemini~2.5 Flash Image & 3.2 & 3.1 & 3.6  \\
\textbf{SceneCraft (Ours)} & \textbf{4.2} & \textbf{4.1} & \textbf{4.4}  \\
\bottomrule
\end{tabular}
\end{table}

\begin{table}[t!]
\centering
\caption{Winning rate (\%) of SceneCraft vs. each baseline (ties excluded), reported per core criterion: Element Composition (EC), Relationship Alignment (RA), and Image Quality (IQ). SceneCraft demonstrates clear advantages across all metrics.}
\vspace{-2mm}
\label{tab:winrate}
\begin{tabular}{lccc}
\toprule
\textbf{Comparison} & \textbf{EC}\(\uparrow\) & \textbf{RA}\(\uparrow\) & \textbf{IQ}\(\uparrow\) \\
\midrule
SceneCraft vs.\ Qwen Image Editing     & 71.0 & 69.8 & 68.5 \\
SceneCraft vs.\ FLUX 1 Kontext     & 77.3 & 75.1 & 74.2 \\
SceneCraft vs.\ Gemini 2.5 Flash Image   & 72.6 & 70.3 & 70.6 \\
\bottomrule
\end{tabular}
\vspace{-3mm}
\end{table}

\begin{figure*}[t!]
\centering
\renewcommand{\arraystretch}{1}
\setlength{\tabcolsep}{0.7pt}
\begin{tabularx}{\textwidth}{>{\centering\arraybackslash}m{0.15\textwidth} >{\centering\arraybackslash}m{0.22\textwidth} *{4}{>{\centering\arraybackslash}m{0.15\textwidth}}}
\textbf{Scene Graph} & \textbf{Operator} & \textbf{Original} & \textbf{Qwen} & \textbf{Flux.1 Kontext} & \textbf{Gemini} \\


\includegraphics[width=\linewidth]{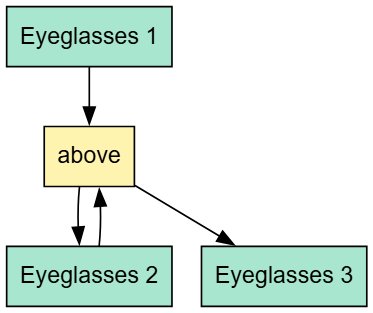}
& {\tiny Add a folded microfiber cleaning cloth under the Eyeglasses 3}
& \includegraphics[width=\linewidth]{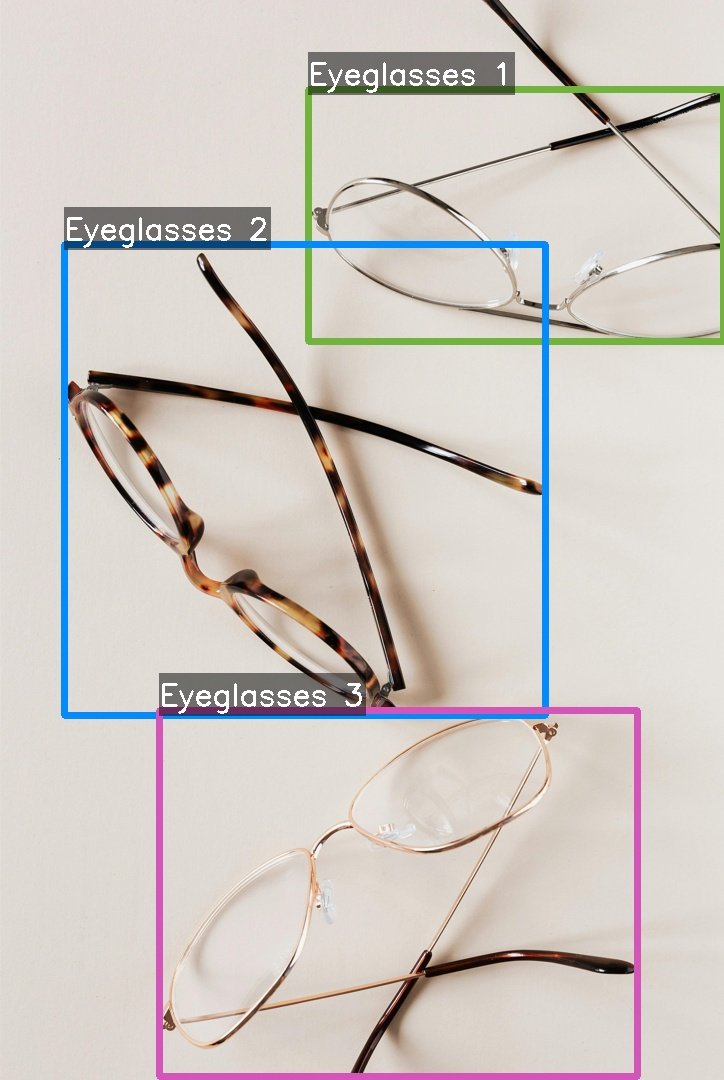} 
& \includegraphics[width=\linewidth]{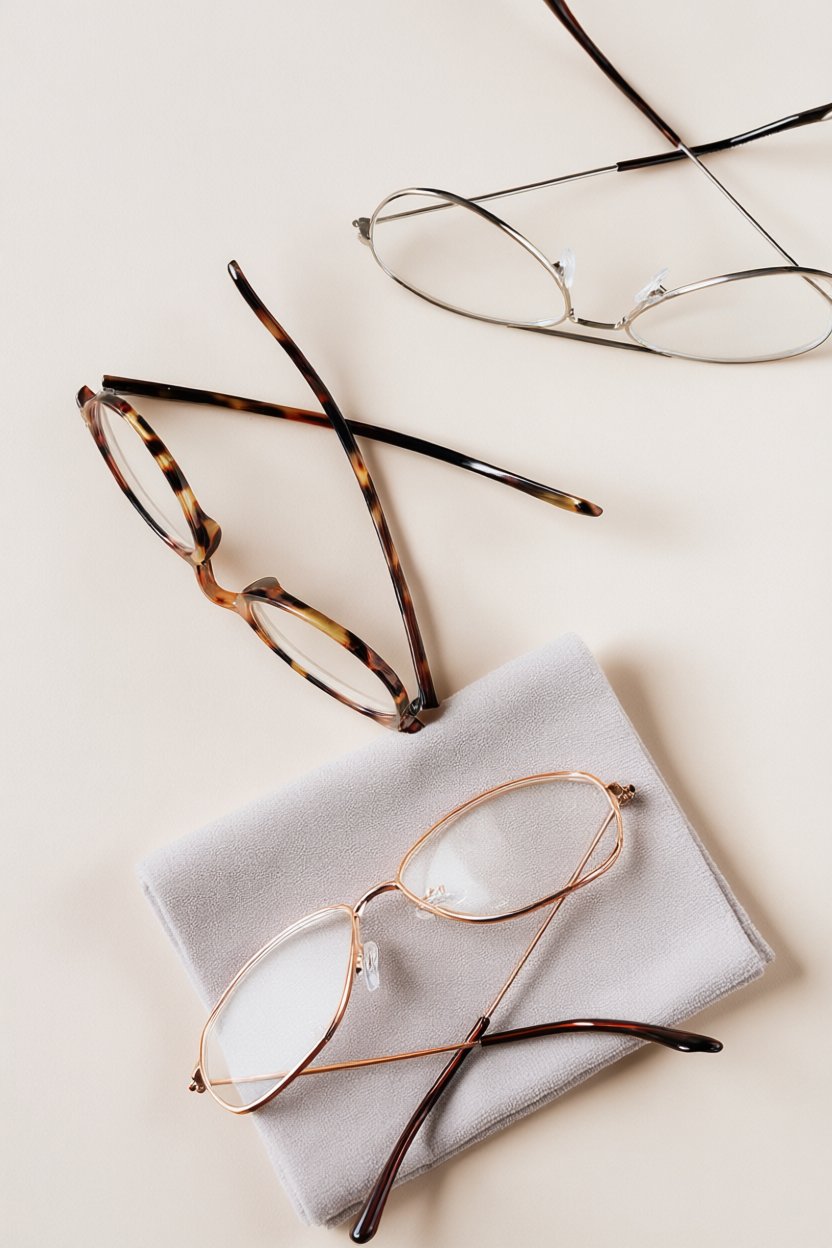} 
& \includegraphics[width=\linewidth]{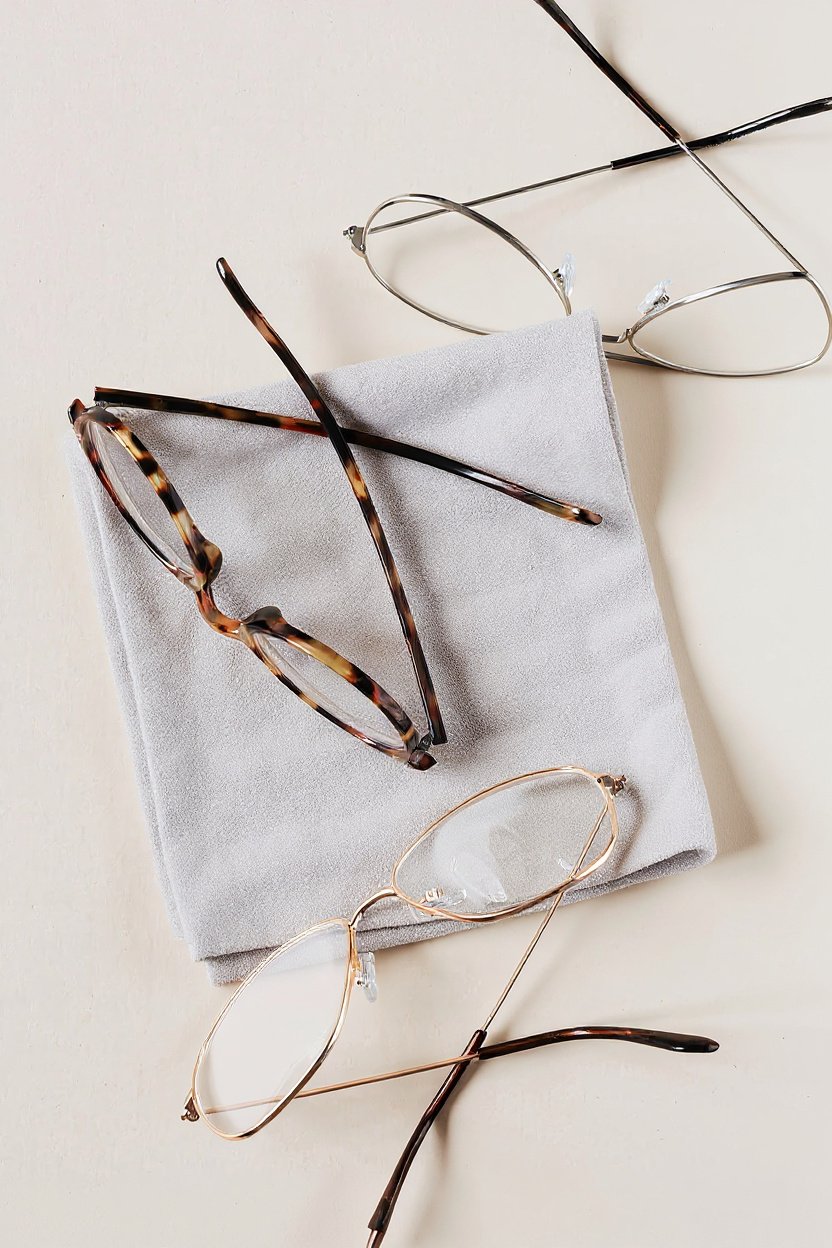} 
& \includegraphics[width=\linewidth]{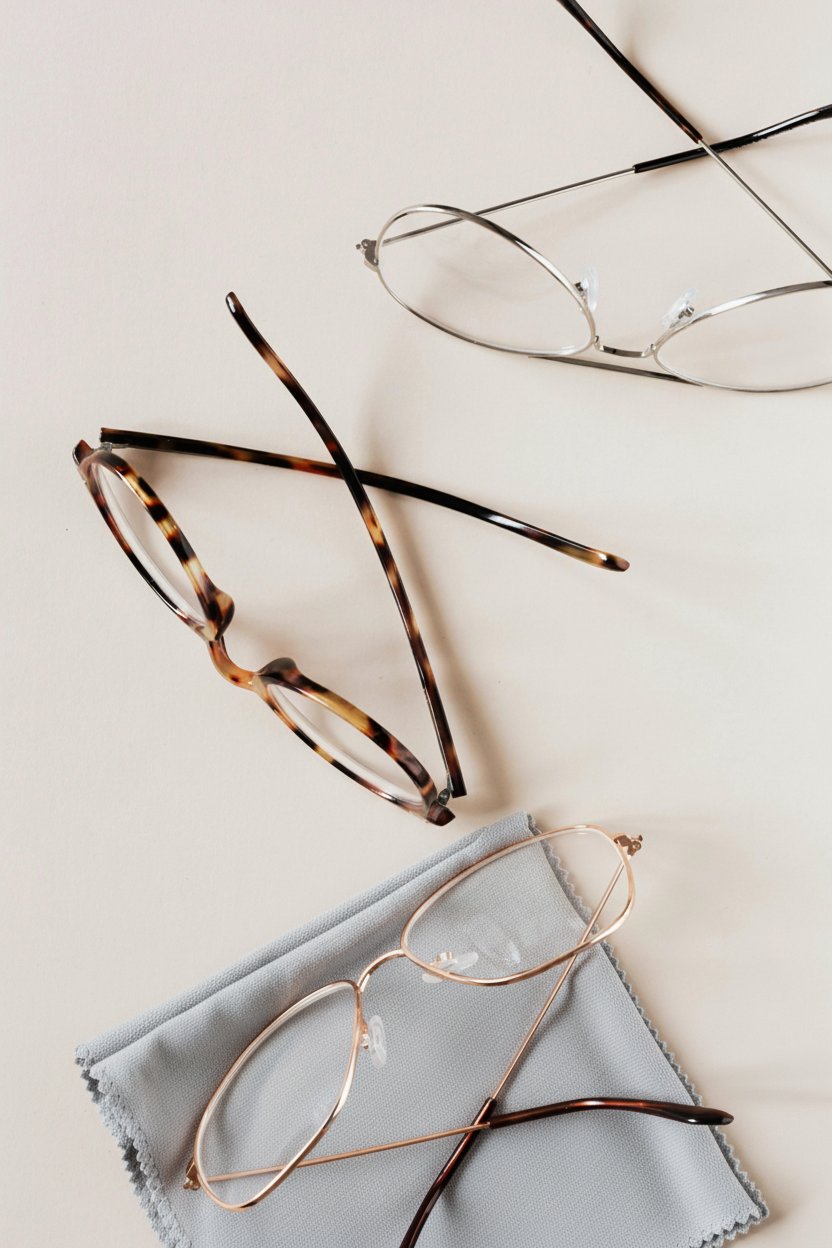} \\


\includegraphics[width=\linewidth]{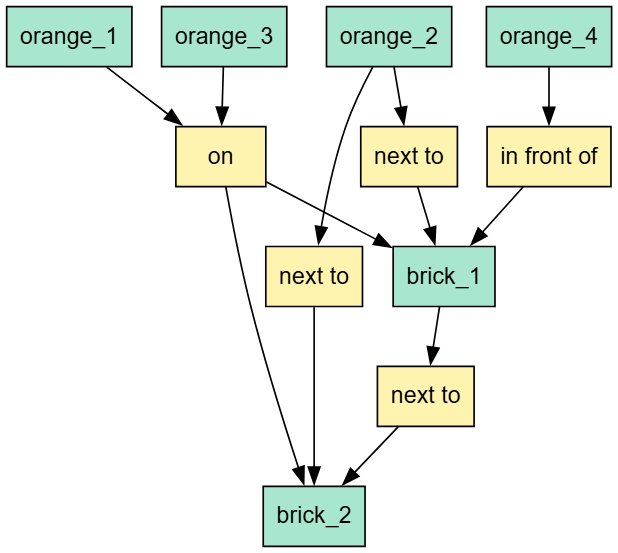}
& {\tiny Turn the Orange 2 into an apple}
& \includegraphics[width=\linewidth]{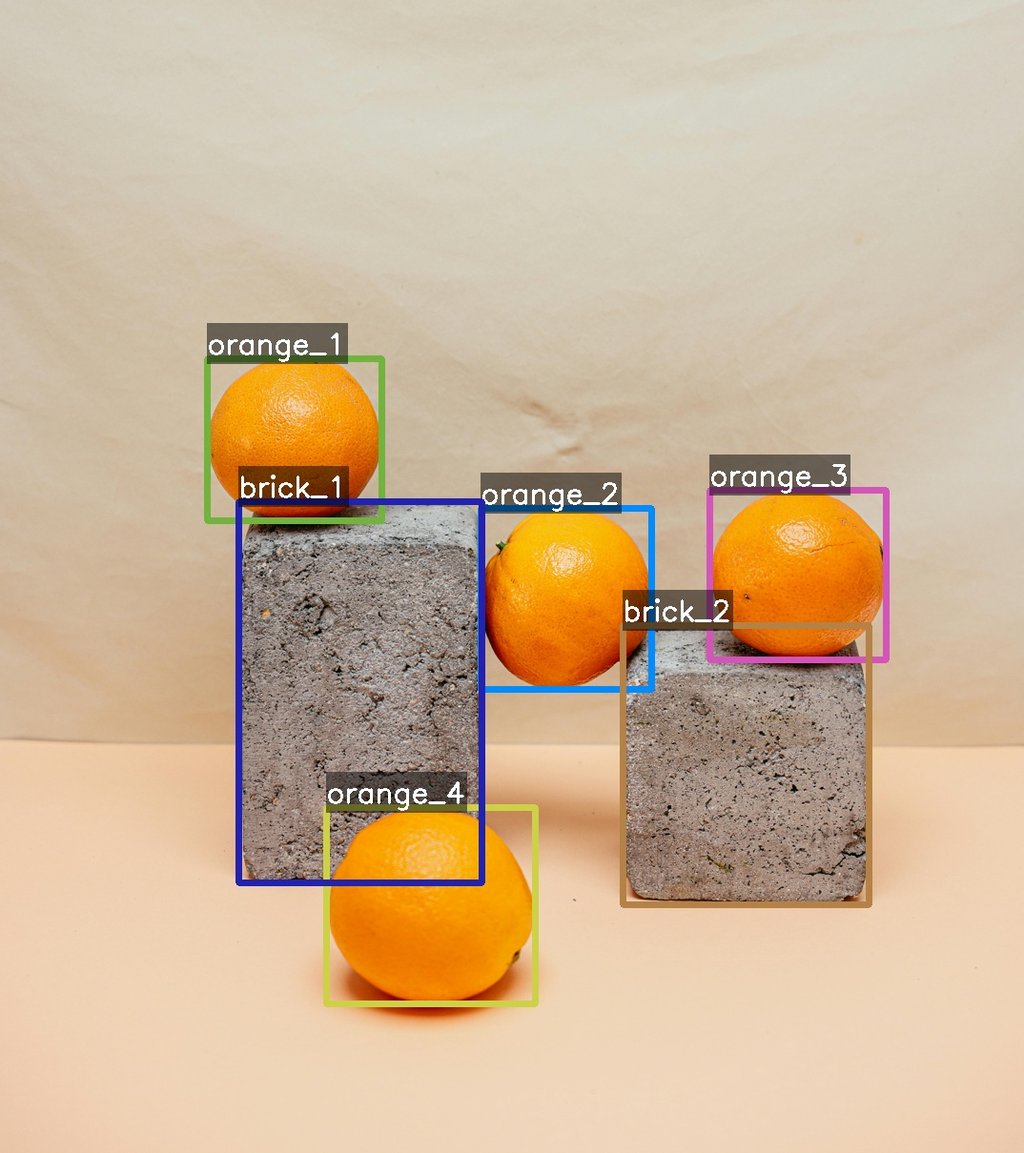} 
& \includegraphics[width=\linewidth]{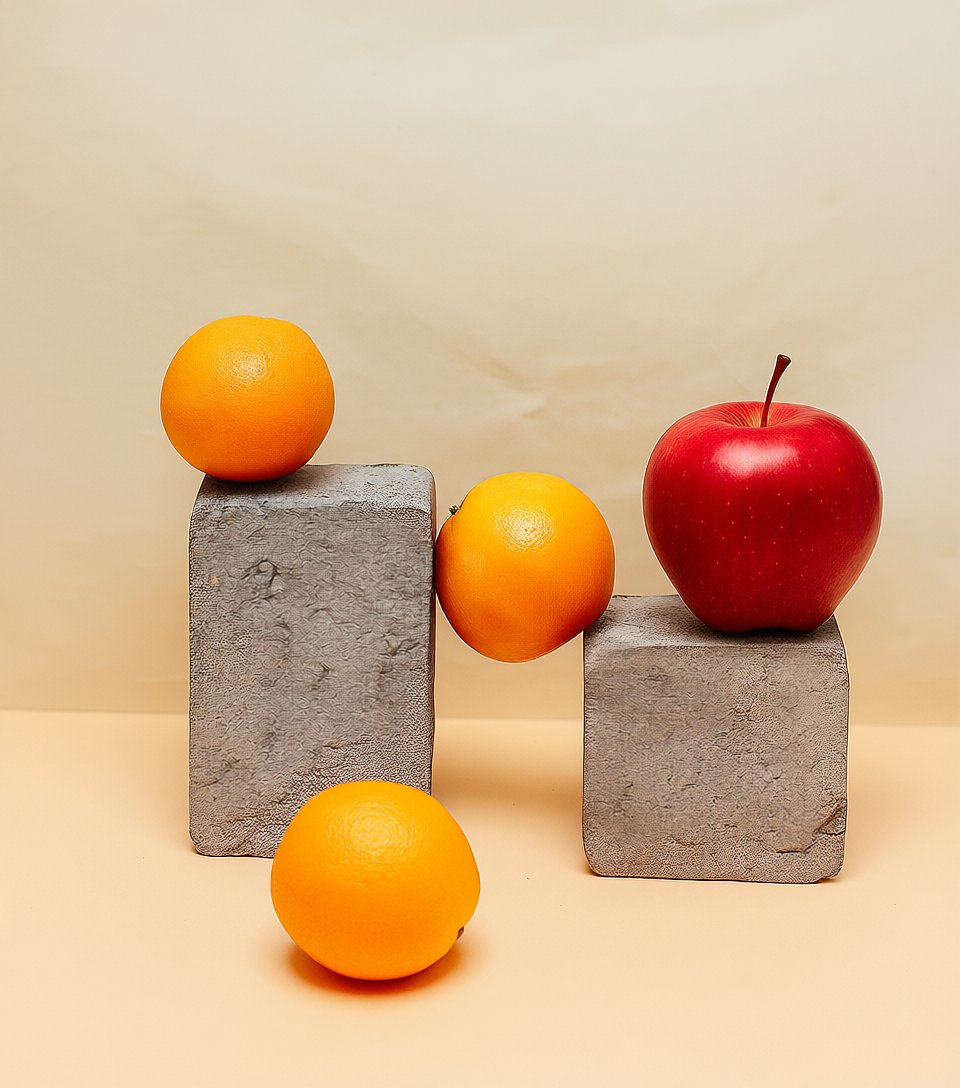} 
& \includegraphics[width=\linewidth]{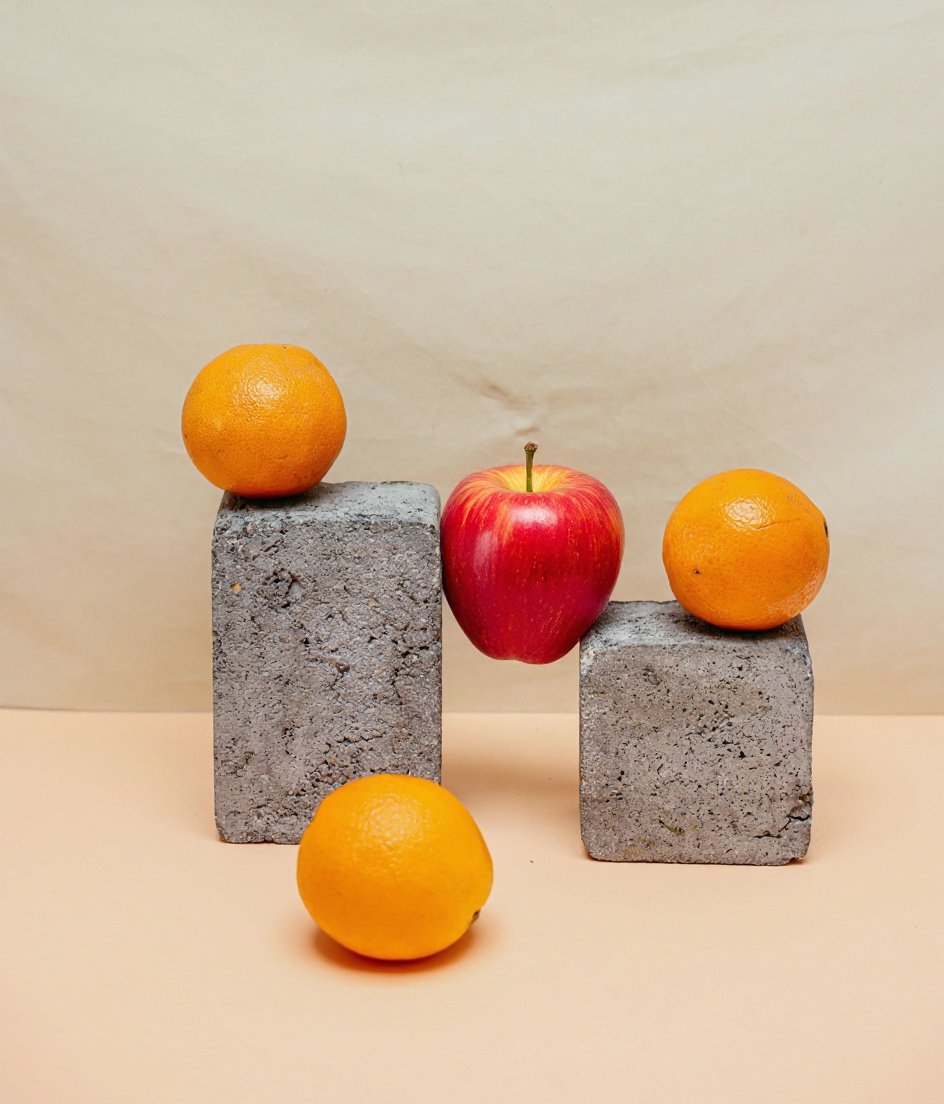} 
& \includegraphics[width=\linewidth]{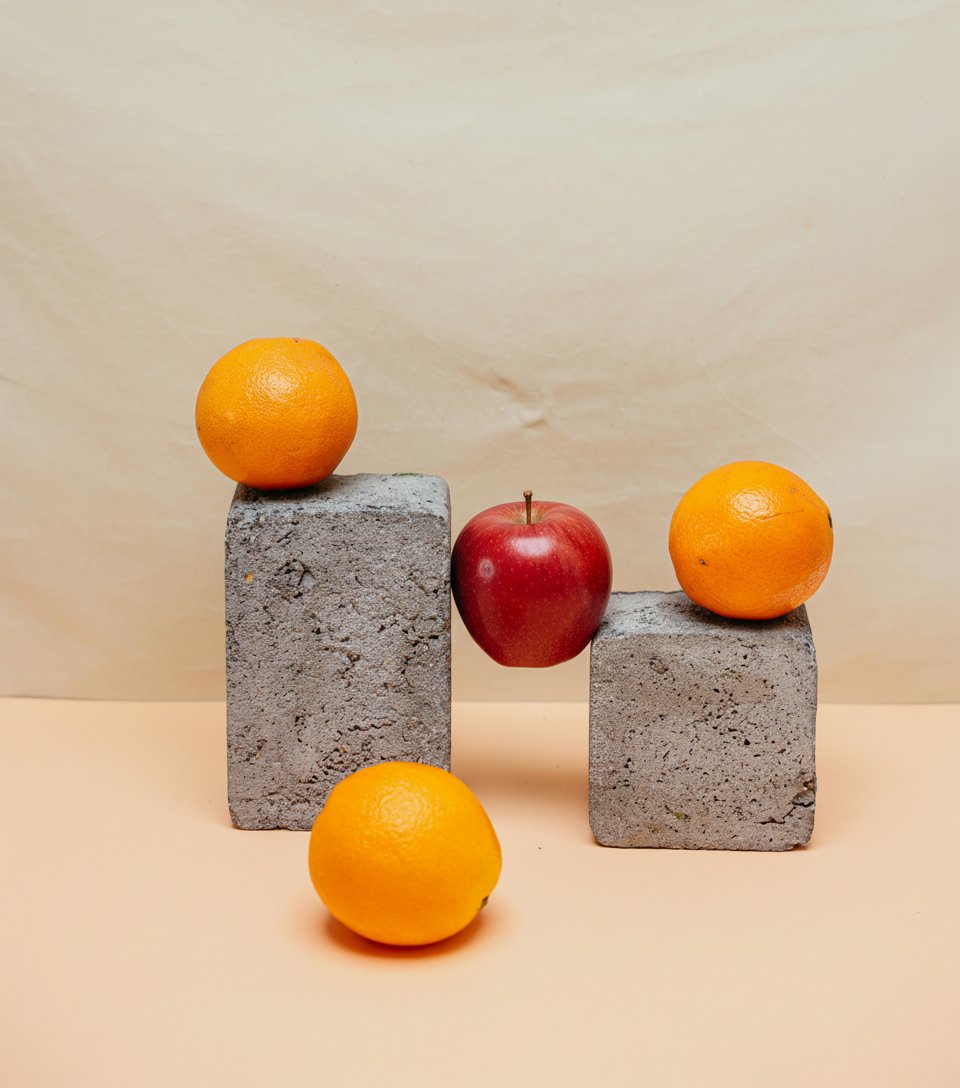} \\

\includegraphics[width=\linewidth]{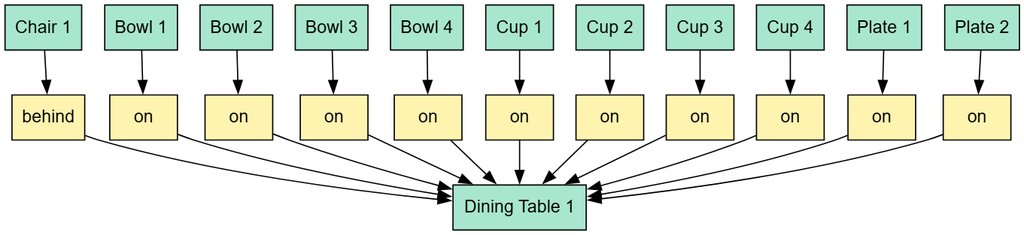}
& {\tiny Add a slice of watermelon on the Bowl 2}
& \includegraphics[width=\linewidth]{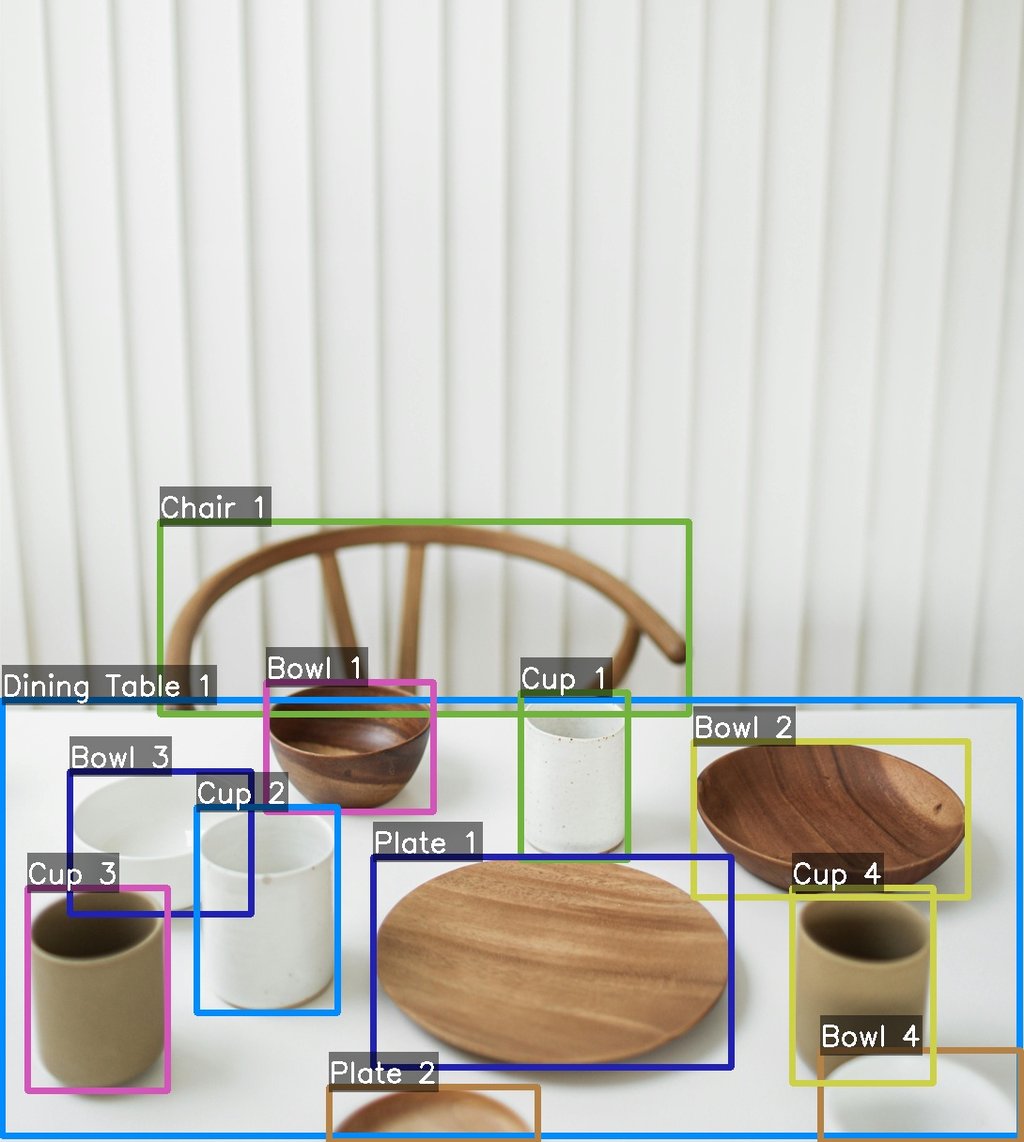} 
& \includegraphics[width=\linewidth]{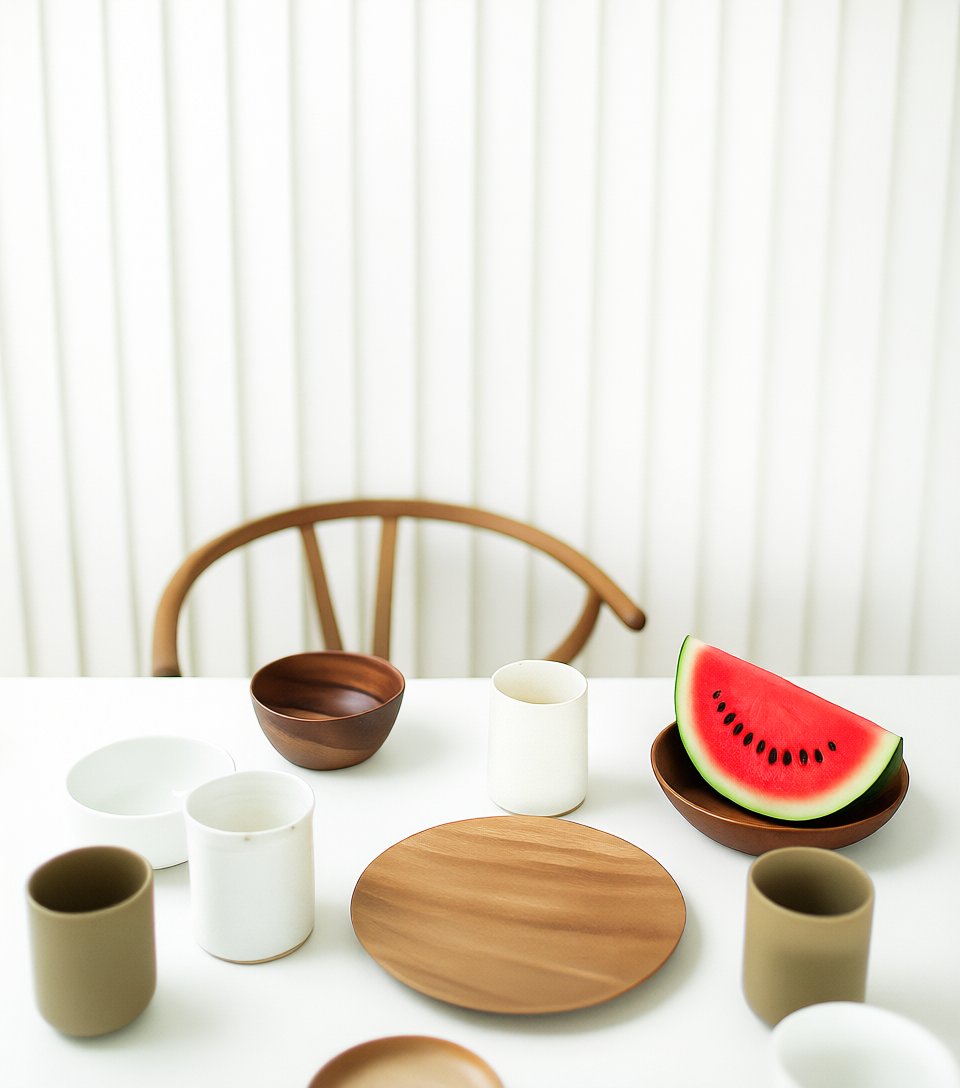} 
& \includegraphics[width=\linewidth]{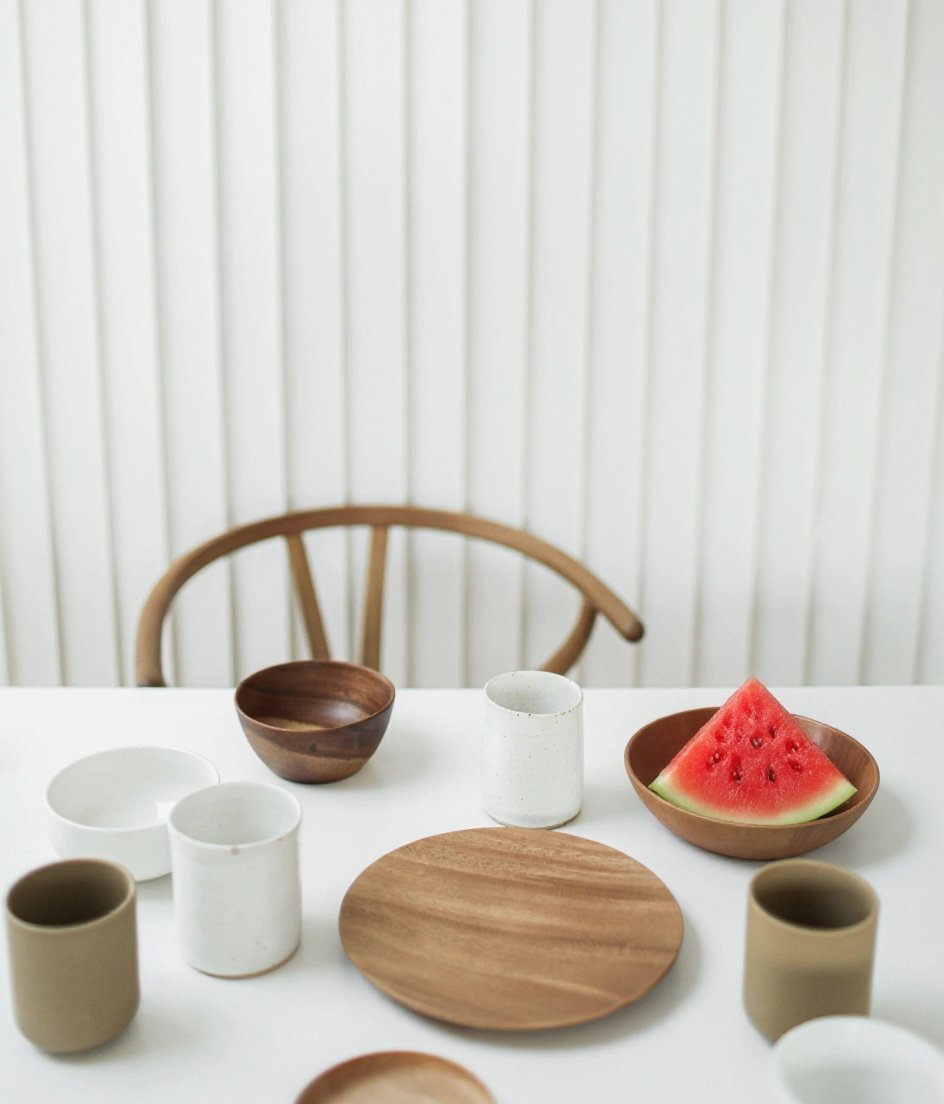} 
& \includegraphics[width=\linewidth]{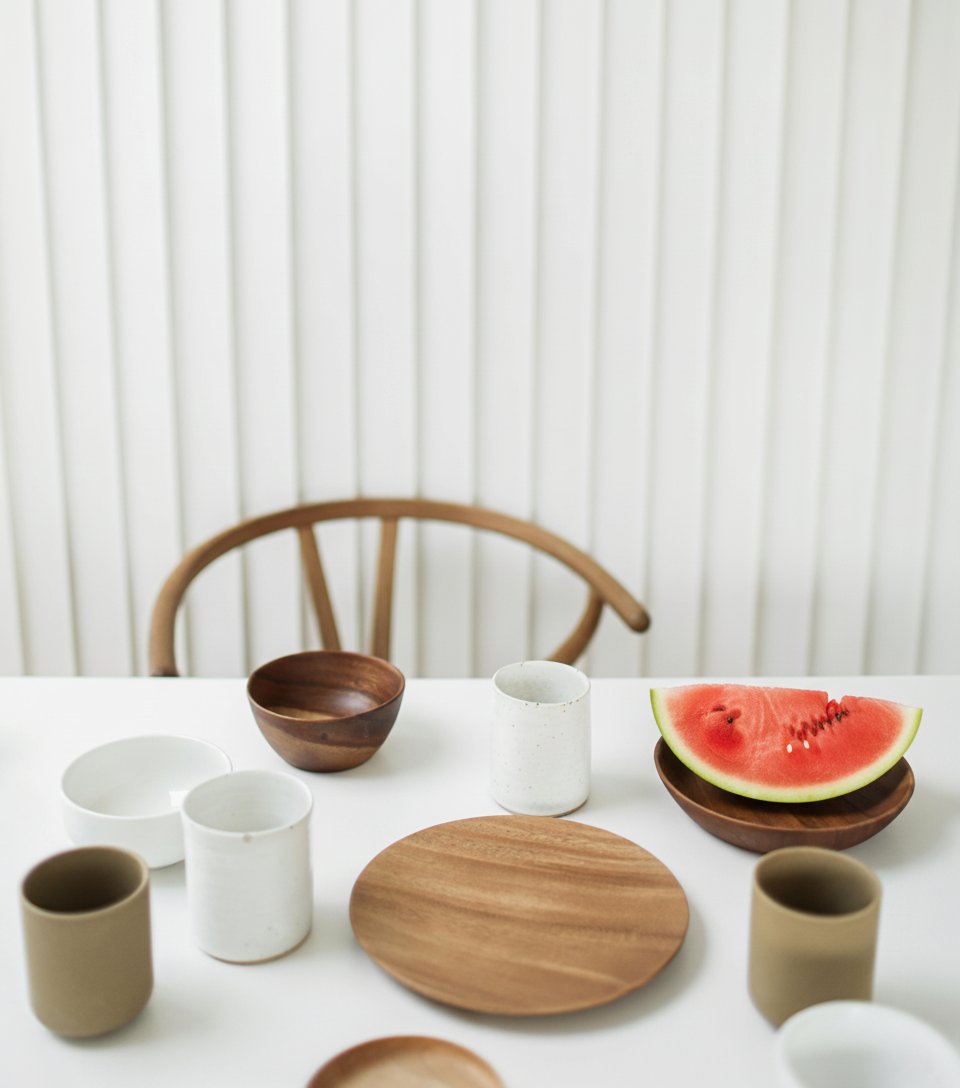} \\

\end{tabularx}
\vspace{-5mm}
\caption{Representative outputs produced by our pipeline across three commercial editors (Qwen Image Editing, FLUX.1 Kontext, Gemini 2.5 Flash Image) for remove, add, and replace operations. The user-manipulated scene graphs (left) act as a reliable bridge to generate structured prompts for the editors.}
\label{fig:pipeline}
\vspace{-3mm}
\end{figure*}

\begin{figure*}[t!]
\centering
\renewcommand{\arraystretch}{1}
\setlength{\tabcolsep}{0.5pt}
\begin{tabularx}{\textwidth}{
  >{\centering\arraybackslash\linespread{0.5}\selectfont}m{0.14\textwidth}  
  >{\centering\arraybackslash}m{0.12\textwidth}  
  *{6}{>{\centering\arraybackslash}m{0.12\textwidth}} 
}
\textbf{\footnotesize Operator}
& 
& \multicolumn{2}{c}{\textbf{\footnotesize Qwen}} 
& \multicolumn{2}{c}{\textbf{\footnotesize Flux.1 Kontext}} 
& \multicolumn{2}{c}{\textbf{\footnotesize Gemini}} \\
& {\footnotesize Original} & {\footnotesize Base} & {\footnotesize Our} & {\footnotesize Base} & {\footnotesize Our} & {\footnotesize Base} & {\footnotesize Our} \\

{\tiny Replace the biggest duck with a dog.}
& \includegraphics[width=\linewidth]{images/photo_3.jpg} 
& \includegraphics[width=\linewidth]{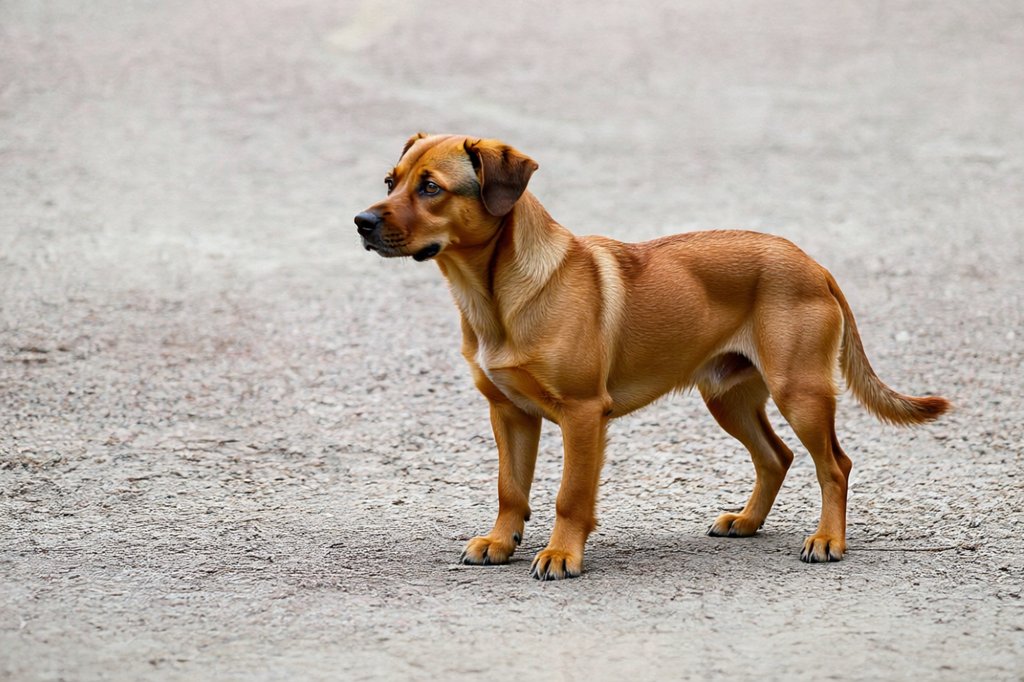} 
& \includegraphics[width=\linewidth]{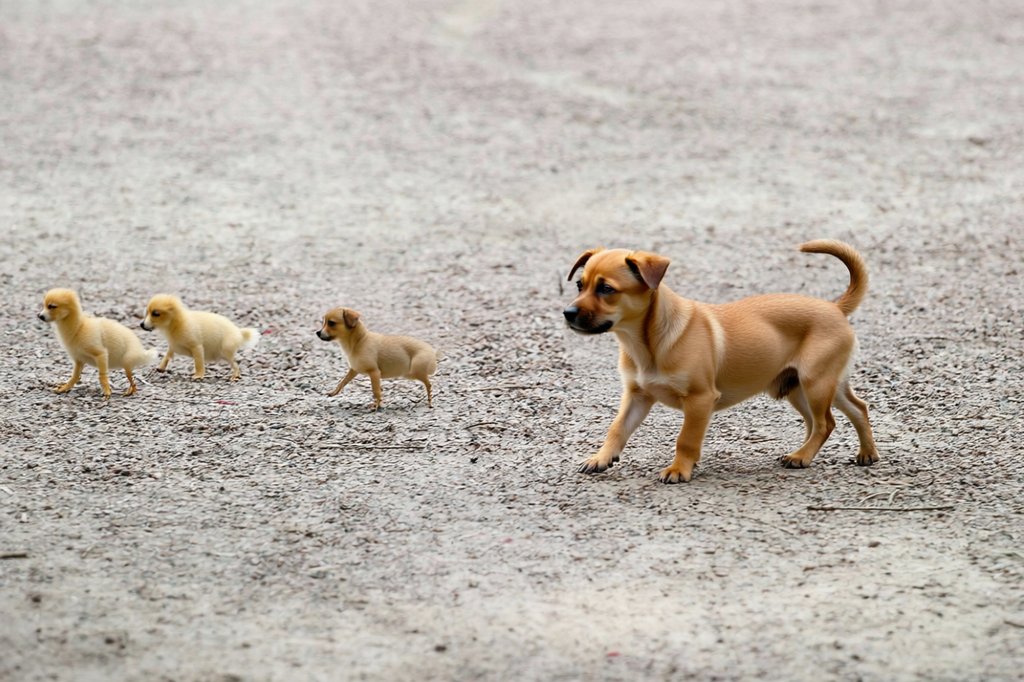} 
& \includegraphics[width=\linewidth]{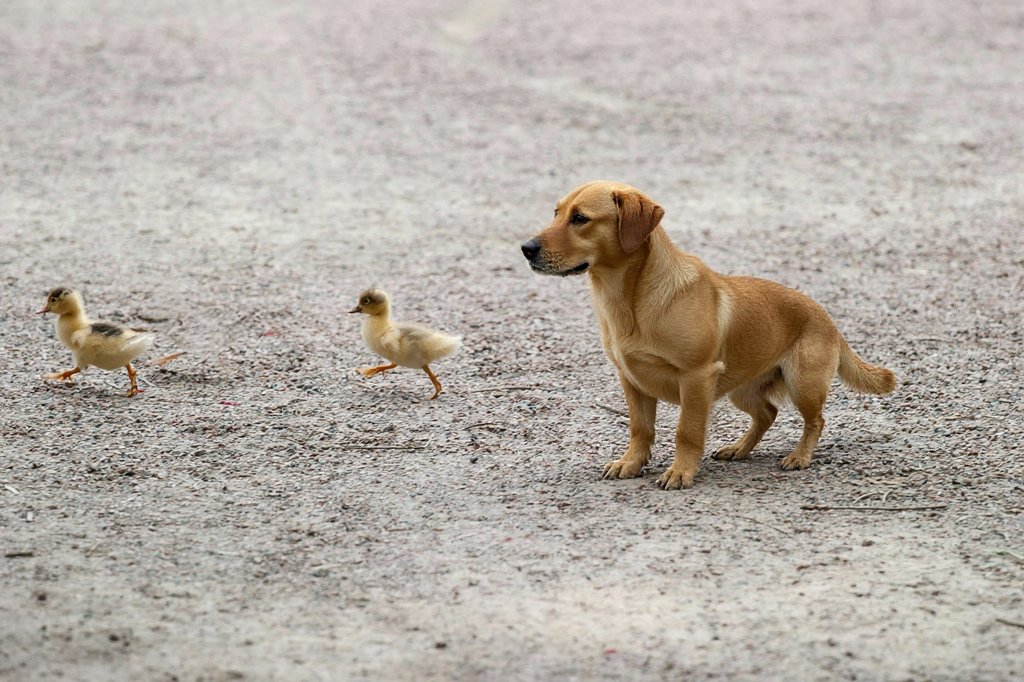} 
& \includegraphics[width=\linewidth]{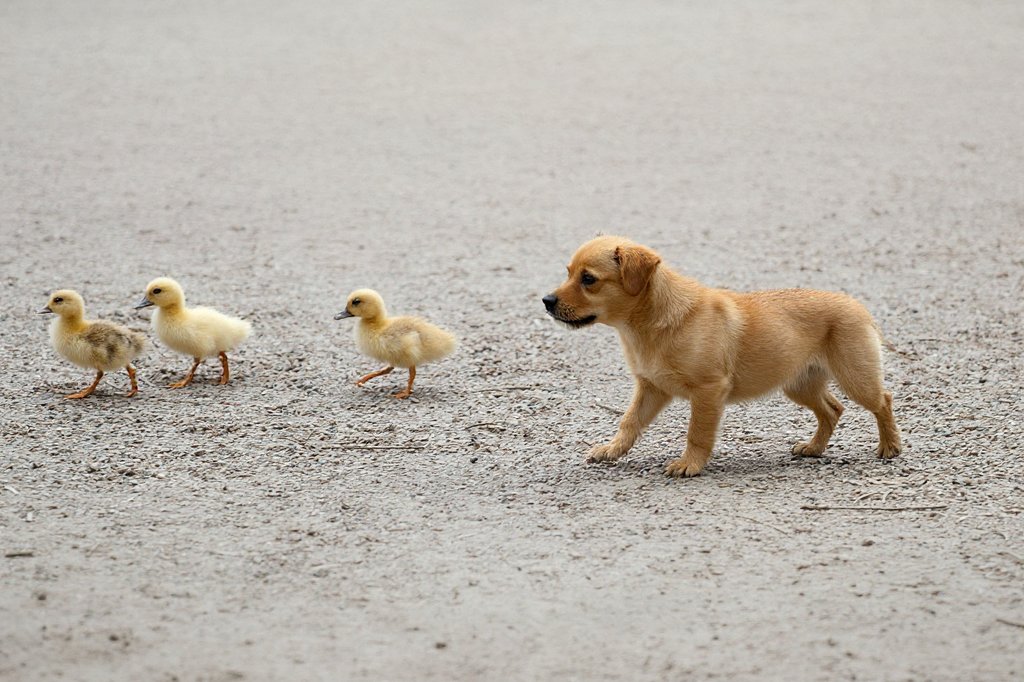} 
& \includegraphics[width=\linewidth]{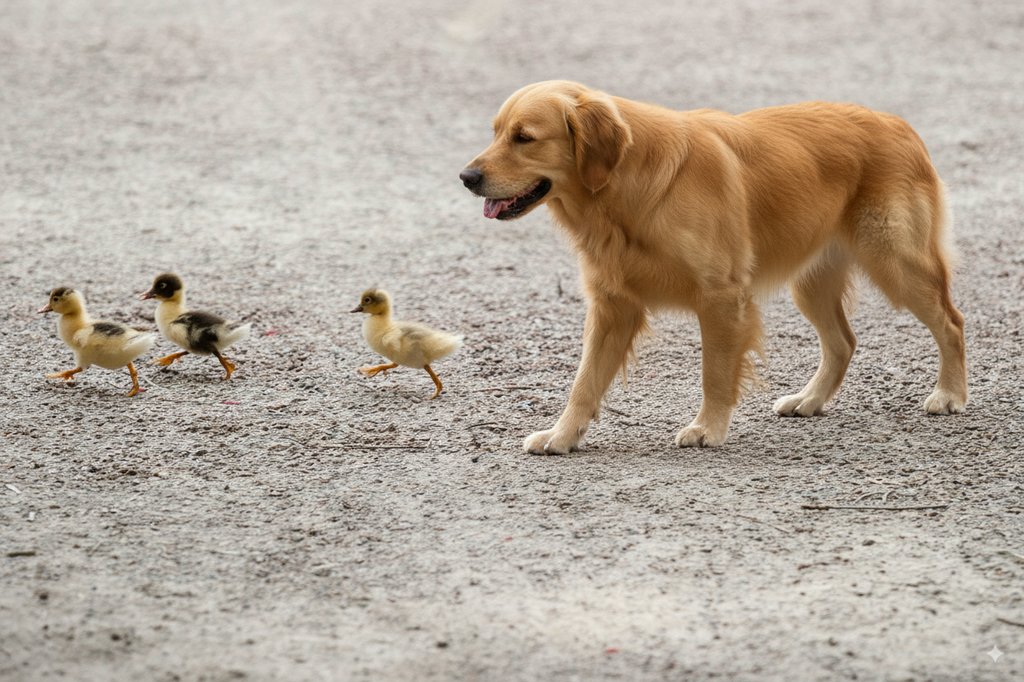}
& \includegraphics[width=\linewidth]{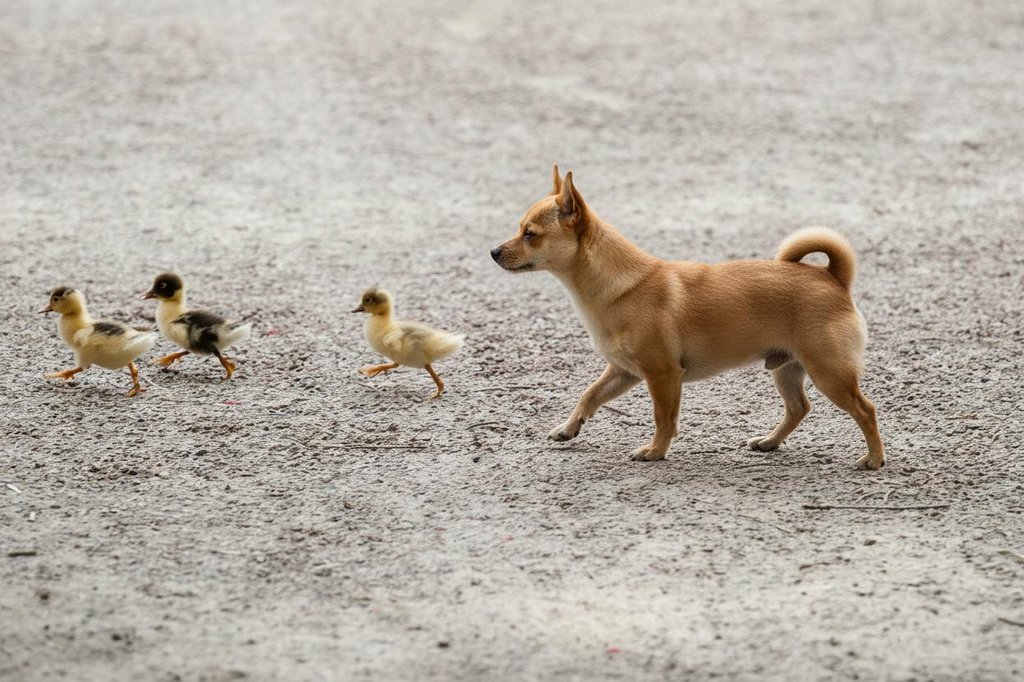} \\

{\tiny Remove the third zebra from the left.}
& \includegraphics[width=\linewidth]{images/photo_5.jpg} 
& \includegraphics[width=\linewidth]{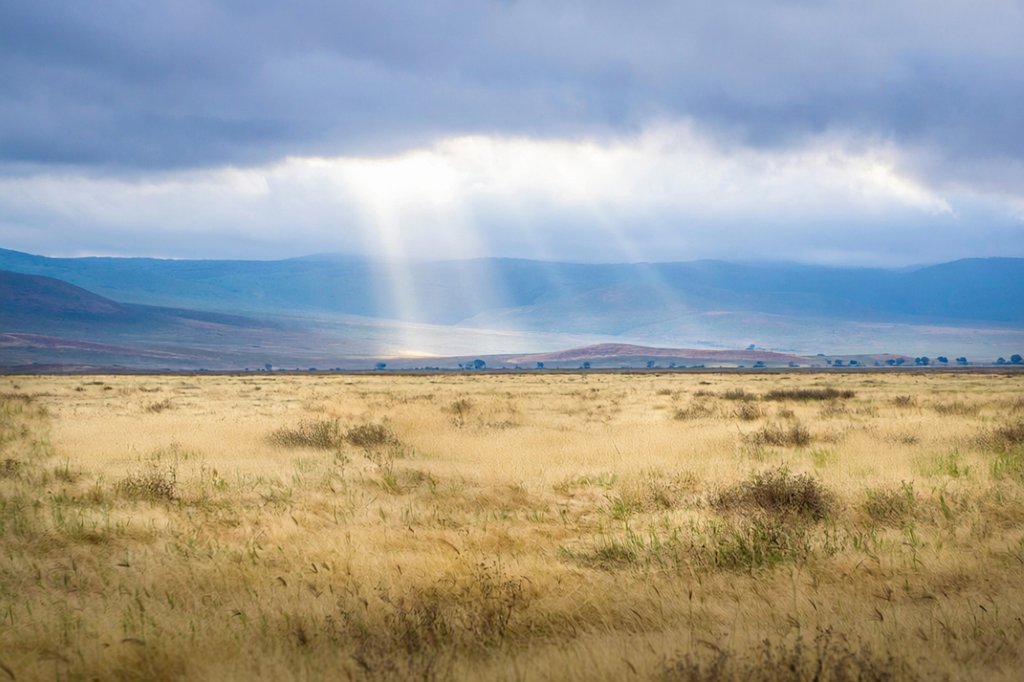} 
& \includegraphics[width=\linewidth]{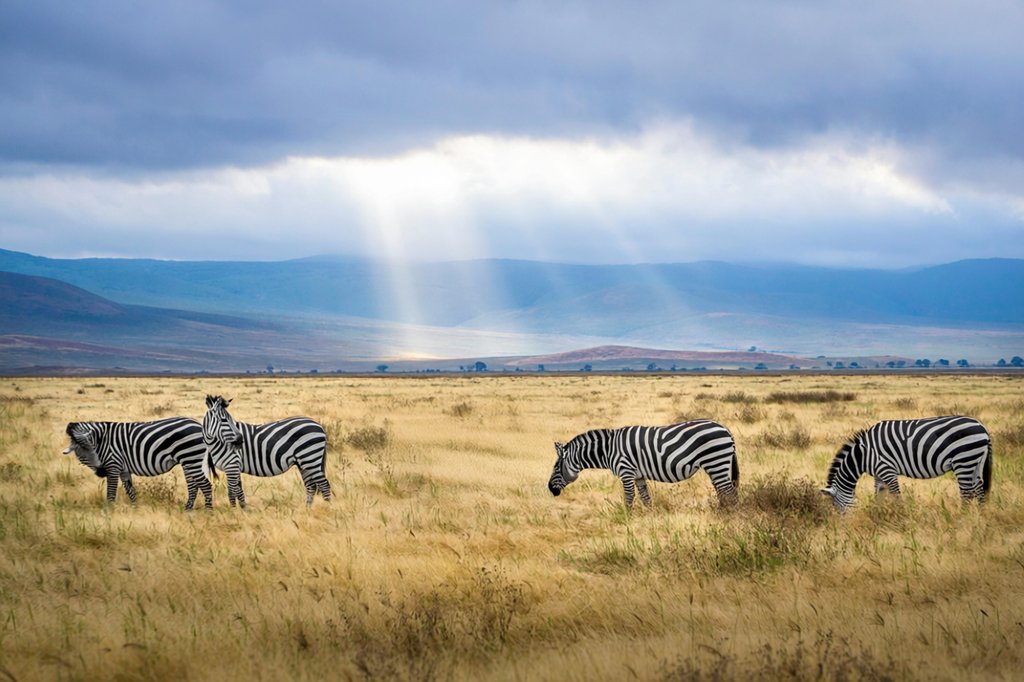} 
& \includegraphics[width=\linewidth]{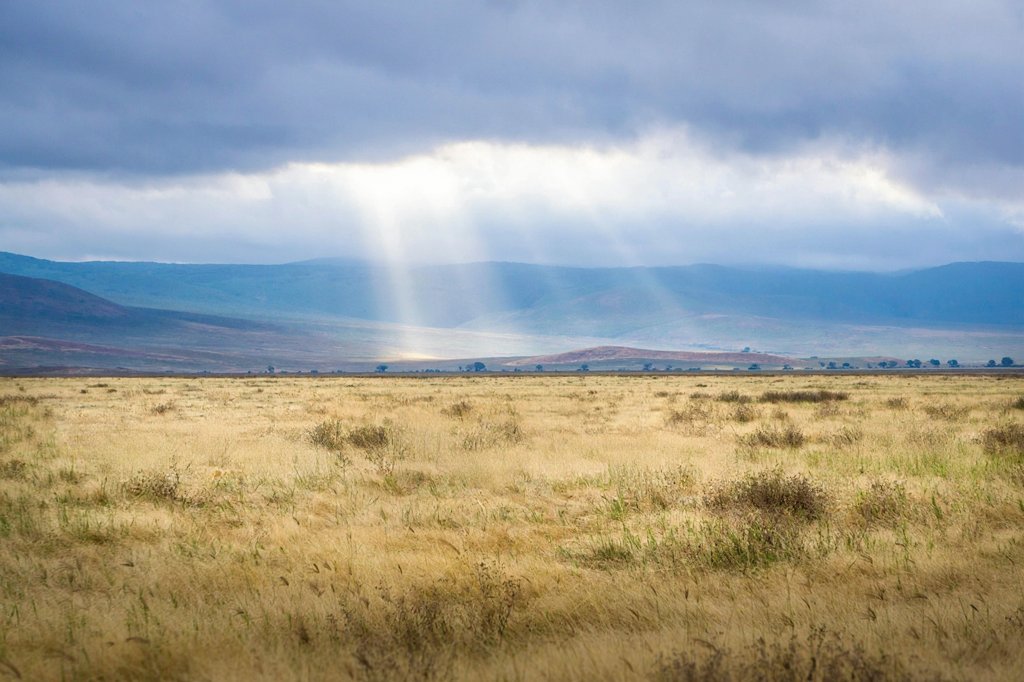} 
& \includegraphics[width=\linewidth]{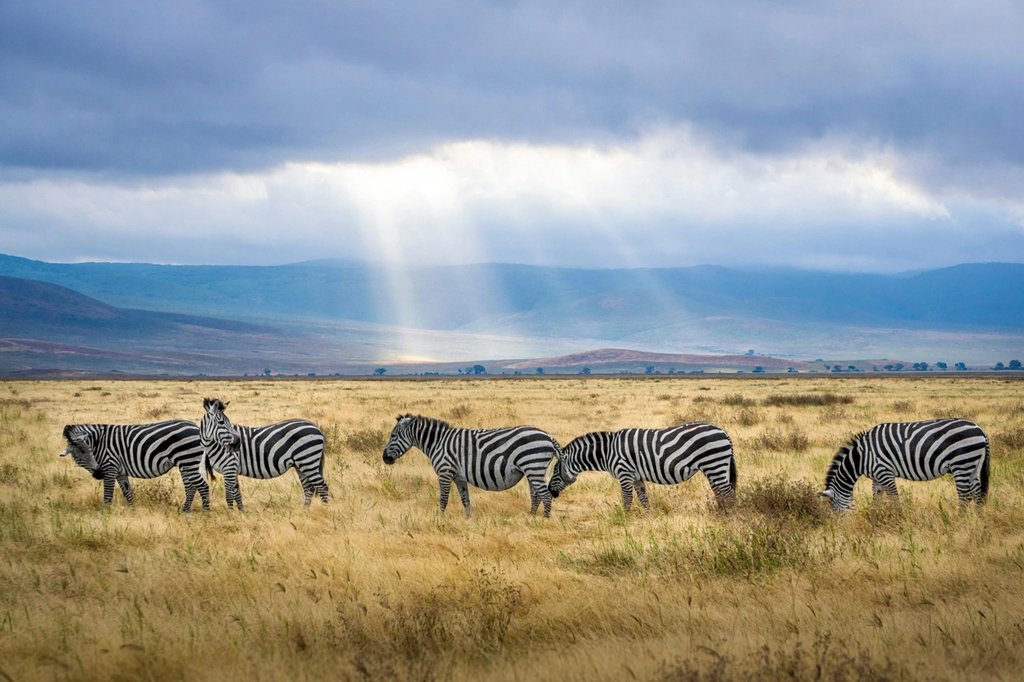} 
& \includegraphics[width=\linewidth]{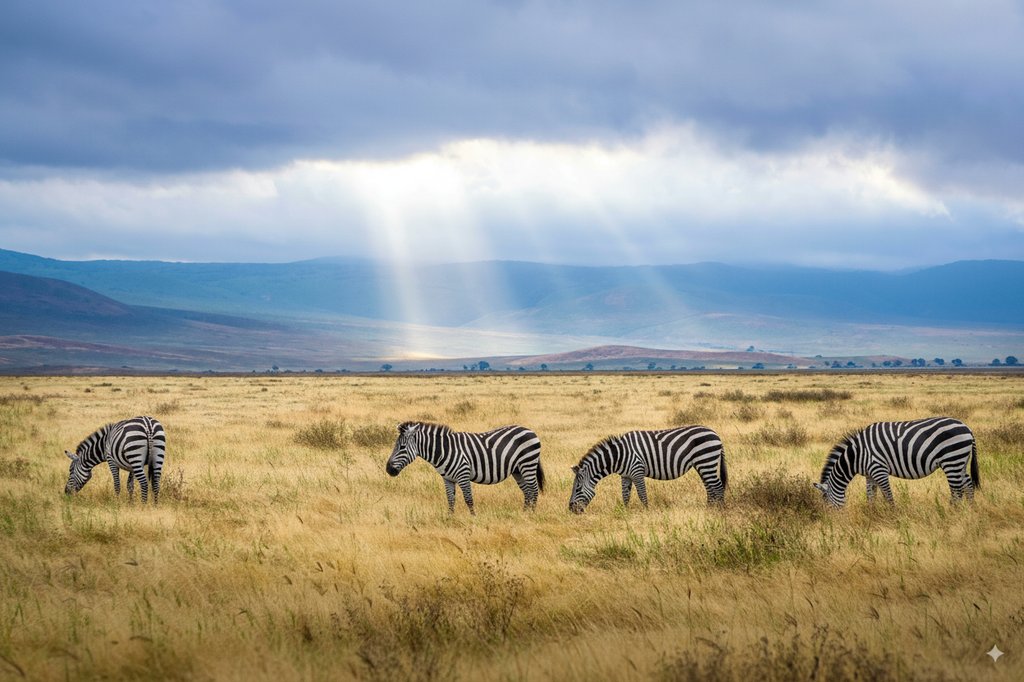}
& \includegraphics[width=\linewidth]{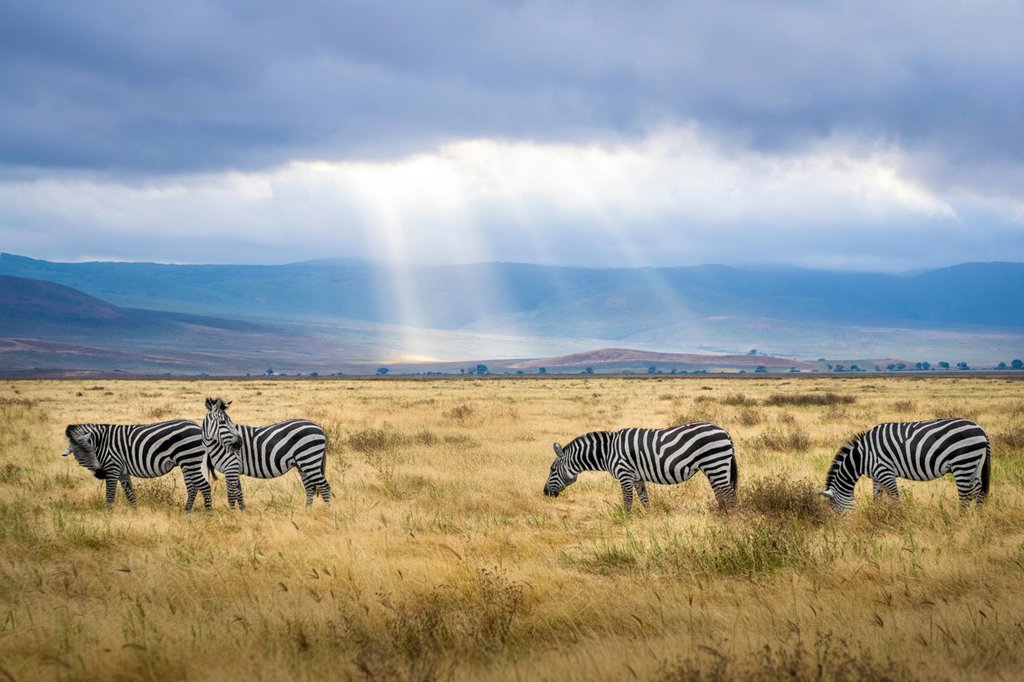} \\

{\tiny Add a ball in front of the second cat from the right.}
& \includegraphics[width=\linewidth]{images/photo_1.jpg} 
& \includegraphics[width=\linewidth]{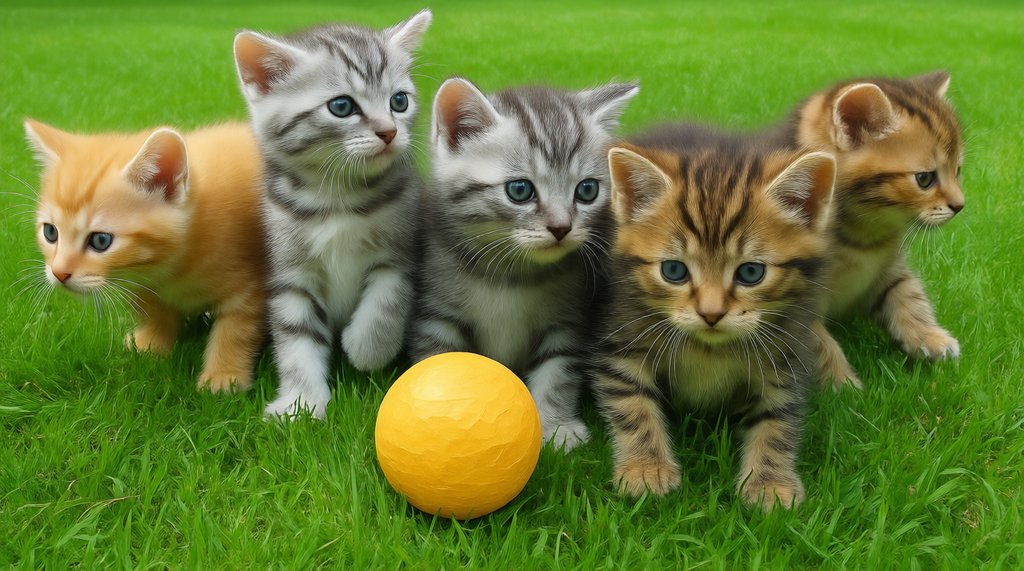} 
& \includegraphics[width=\linewidth]{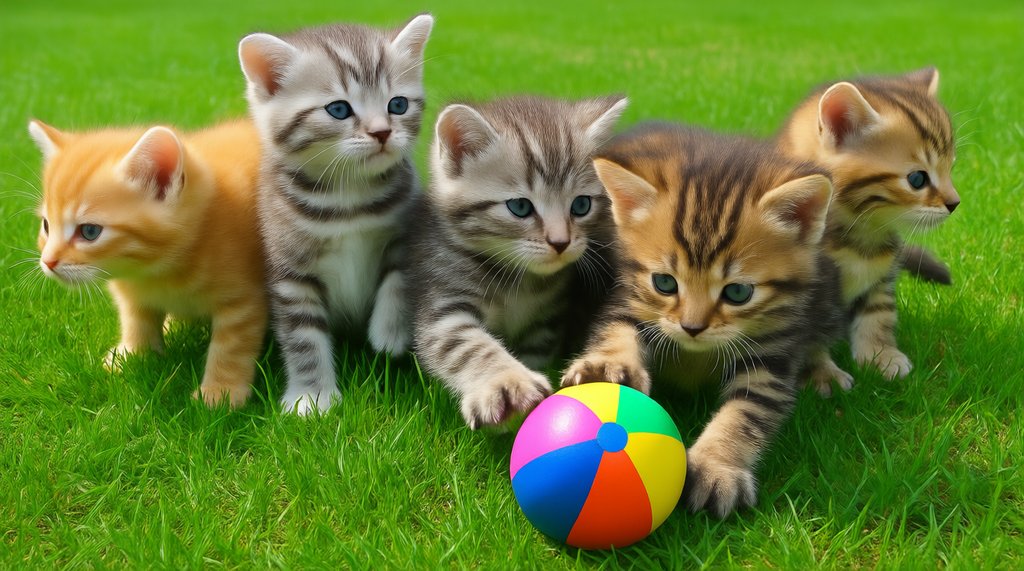} 
& \includegraphics[width=\linewidth]{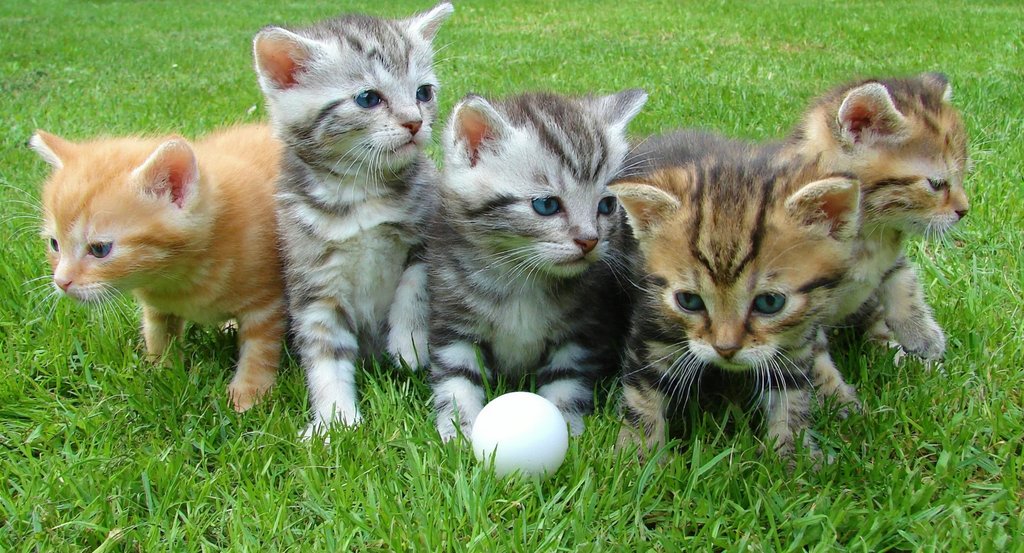} 
& \includegraphics[width=\linewidth]{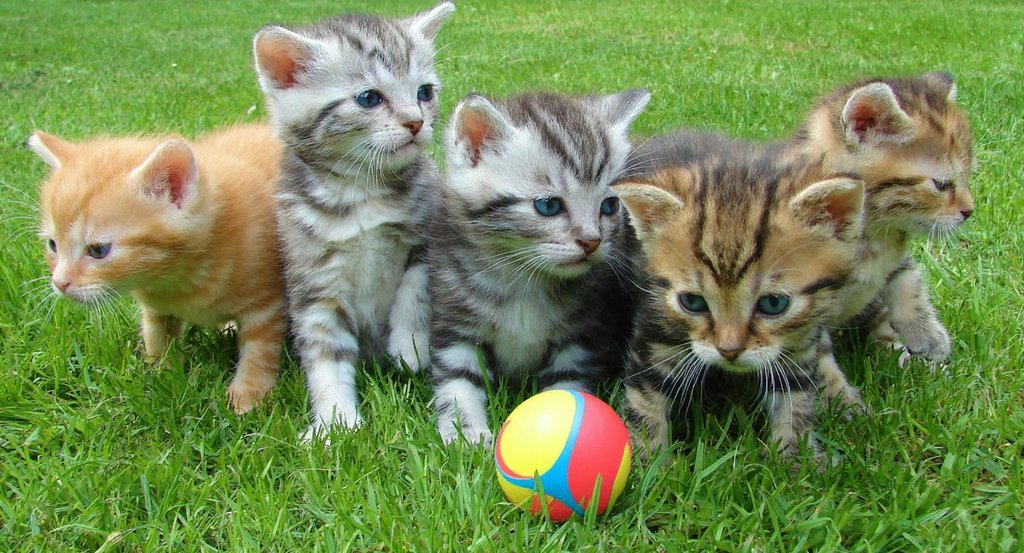} 
& \includegraphics[width=\linewidth]{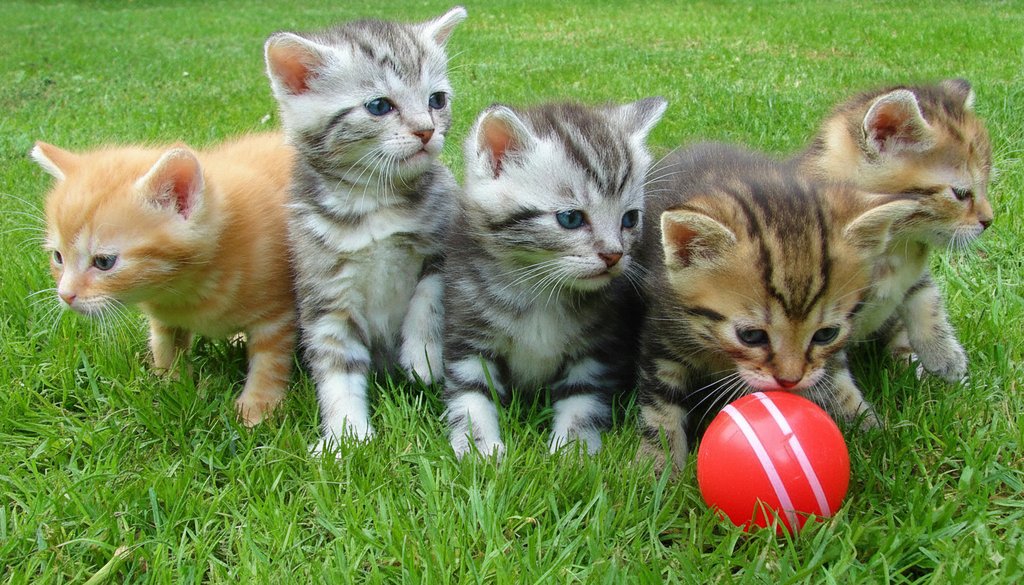}
& \includegraphics[width=\linewidth]{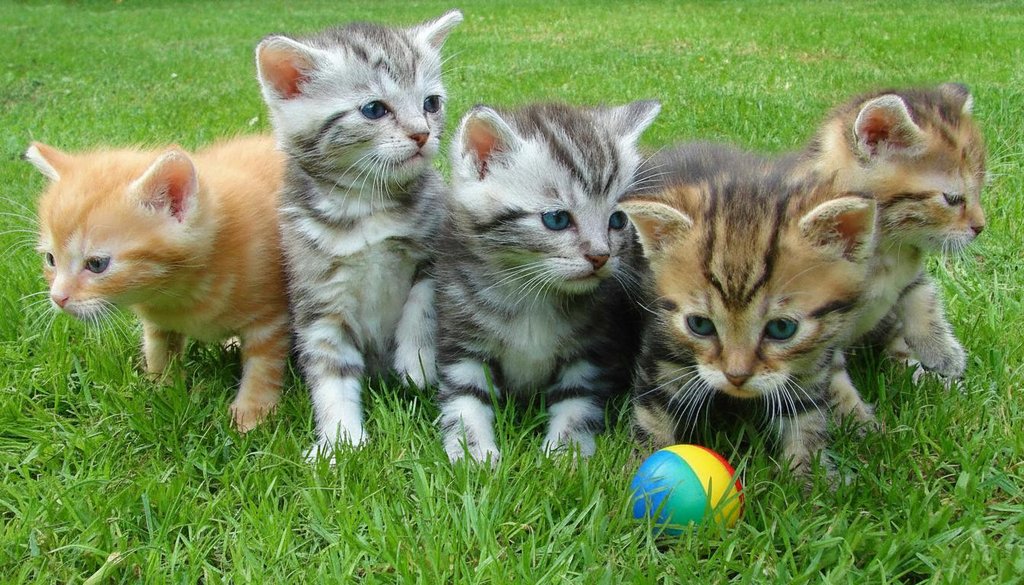} \\

\end{tabularx}
\vspace{-5mm}
\caption{Qualitative comparison of raw-prompt baselines (Base) vs. our graph-enhanced prompting (Our) for the same generative editors (Qwen Image Editing, FLUX.1 Kontext, Gemini 2.5 Flash Image). SceneCraft significantly improves precise object placement and background consistency.}
\label{fig:qual}
\vspace{-3mm}
\end{figure*}

\subsubsection{Qualitative Results}

Fig.~\ref{fig:pipeline} visualizes representative outputs produced through our pipeline across the three backbones. Fig.~\ref{fig:qual} compares raw-prompt baselines ({Base}) to our graph-enhanced prompting ({Our}) for the same backbones. This side-by-side comparison  highlights that graph-enhanced prompting reduces semantic drift and improves background preservation versus raw prompting.

\subsection{System Performance Evaluation}

To deeply understand how SceneCraft's interface design impacts the creator's workflow, we conducted a usability study to evaluate the system's ease of use, UI friendliness, and impact on user cognitive load compared to standard commercial chat-based interfaces, such as ChatGPT and Qwen Chat. After participants used our system and gave technical evaluation scores, we collected both qualitative feedback via semi-structured interviews and quantitative data to evaluate the user experience comprehensively.

\begin{table}[t!]
\centering
\caption{System performance evaluation, using Mean Opinion Scores (MOS; 1–5 scale). Participants rated SceneCraft highly on Ease of Use and UI Friendliness compared to traditional chat-based LLM interfaces.}
\label{tab:system_performance}
\vspace{-3mm}
\begin{tabular}{lcc}
\toprule
\textbf{System} & \textbf{Ease of Use}\(\uparrow\) & \textbf{UI Friendliness}\(\uparrow\) \\
\midrule
ChatGPT & 3.50 & 4.25 \\
Qwen Chat & 3.45 & 4.25 \\
\textbf{SceneCraft (Ours)} & \textbf{3.90} & \textbf{4.40}  \\
\bottomrule
\end{tabular}
\end{table}

\textbf{System Usability and Learnability:} The 5-point Likert scale analysis in Table~\ref{tab:system_performance} revealed that SceneCraft achieved a high average score, indicating that participants considered the tool to have excellent usability and a gentle learning curve. Furthermore, results from the Post-Study System Usability Questionnaire (PSSUQ) highlighted strong overall satisfaction with the interface design, with participants particularly valuing the system's ability to help them effectively complete tasks and its visually pleasant layout.

\begin{figure}[t]
  \centering
  \includegraphics[width=\columnwidth]{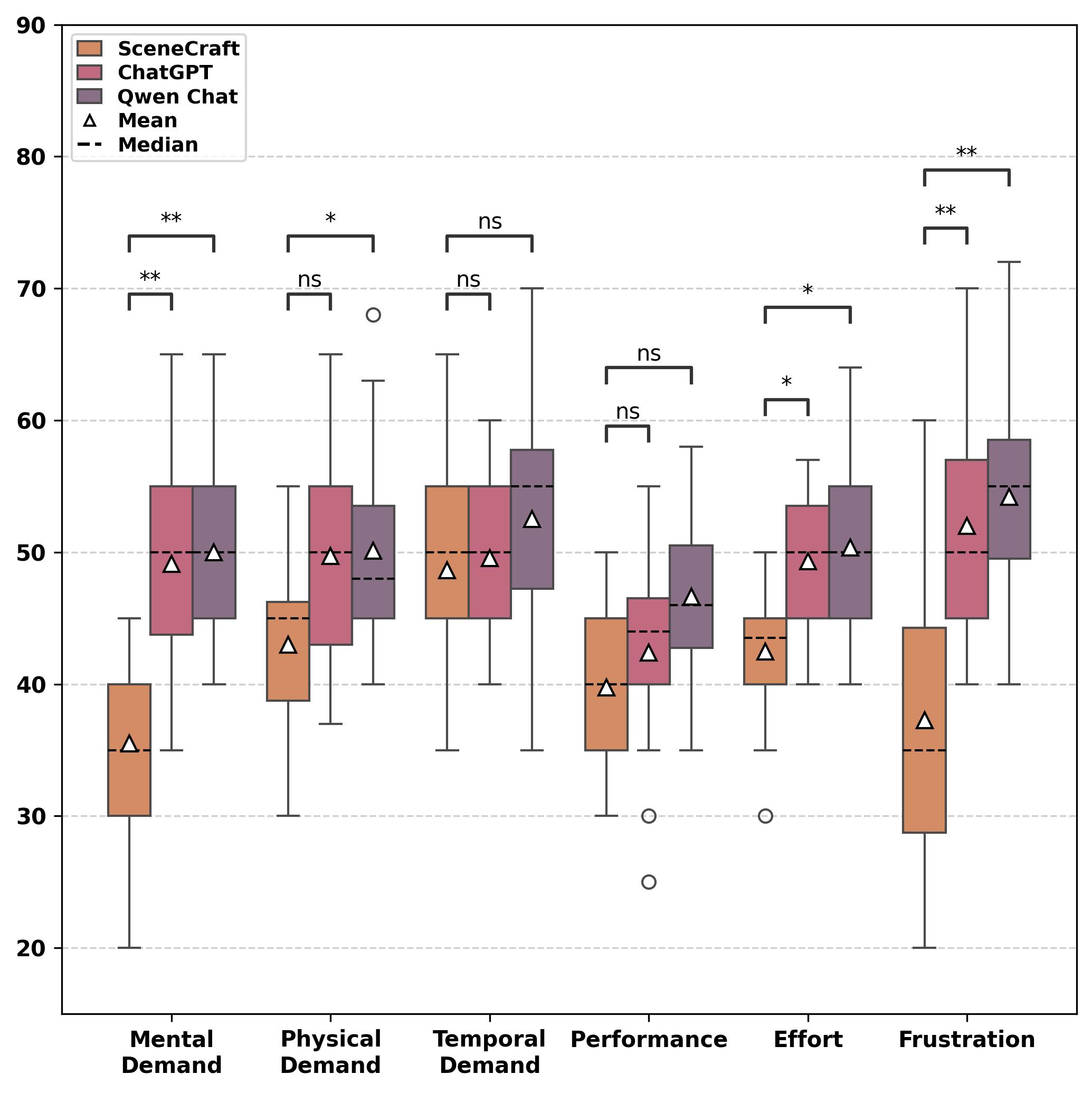}
  \caption{Task load results measured by the NASA-TLX from the User Study (lower scores are better). The $y$-axis displays the six NASA-TLX subscales, and the $x$-axis shows the corresponding scores. SceneCraft significantly lowers the cognitive burden, frustration, and mental demand associated with complex image editing compared to commercial chat-based image editing interfaces. (* indicates $p$ < 0.05, ** indicates $p$ < 0.01)}
  \label{plot:tlx}
\end{figure}

\textbf{Cognitive Load Reduction:} We utilized the NASA Task Load Index (NASA-TLX) to assess the perceived workload associated with crafting edits (Fig.~\ref{plot:tlx}). For the statistical analysis, we used Wilcoxon signed-rank tests~\cite{woolson2007wilcoxon} to analyze the data, with statistical significance $p < 0.05$. In the raw-prompt baseline condition using commercial tools, participants reported high frustration due to the unpredictability of the generative models. SceneCraft significantly lowered this; by translating ambiguous text formulation into direct visual graph manipulation, the system effectively reduced the cognitive burden required to achieve complex spatial edits.

\textbf{Qualitative Feedback:} Our thematic analysis of the post-study interviews highlighted several ways SceneCraft’s UI shifted participants' creative strategies away from frustrating trial-and-error loops.
\begin{itemize}
    \item \textit{Intuitive Operations for Non-Experts:} Beginners without professional backgrounds indicated that the Scene Graph Editor was highly intuitive. They noted that even without prompt engineering skills, they could complete complex structural designs through simple node manipulations, giving them a high sense of achievement. Experienced creators similarly praised the low learning curve, noting that they could start creating with almost no onboarding cost.

    \item \textit{System Responsiveness and Visual Clarity:} Participants commended the system’s intuitive design and responsive interface. Users described the graph interactions as having ``very good responsiveness, with operations feeling smooth and natural,'' noting that the interface design clearly separates functional modules, making the entire editing pipeline easy to comprehend.

    \item \textit{Transparency and Control:} A recurring theme was the frustration users face with the ``black box'' nature of standard text-to-image models. Participants emphasized that the structured layout of SceneCraft effectively linked their inputs directly to analytical outputs, significantly reducing the time spent debugging why an edit failed. By visualizing the exact relationships the model was prioritizing, SceneCraft provided a ``clear cognitive map'' that gave users greater agency over their final outputs.
\end{itemize}

\subsection{Insights}

SceneCraft’s gains are most pronounced in EC and RA for \emph{Remove/Add}, where preserving background and enforcing relations are critical. For \emph{Replace}, it maintains pose/lighting better than raw-prompt baselines, improving IQ without sacrificing EC/RA. We attribute these improvements to: (1) scene-graph-driven prompt construction that removes linguistic ambiguity, and (2) multi-backbone aggregation that mitigates single-model failure modes. Indeed, qualitative inspections confirmed that enriched context improves object placement and relationship preservation, while aggregation ensures higher success rates across diverse edits:
\begin{itemize}
    \item \textit{Without enriched scene graph context:} Using only object categories without structured relationships led to misaligned edits. The UI's relational links are essential for guiding the backbone models.
    
    \item \textit{No candidate selection:} Showing only one edited result notably reduced user satisfaction compared to offering a gallery of multiple options, proving that diverse generation (DG3) is a vital feature of the user experience
\end{itemize}






\section{Discussions}

SceneCraft demonstrates that explicit semantic structures can bridge the gap between user intent and generative model execution. By manipulating a scene graph, users are no longer required to reverse-engineer how an LLM interprets spatial prepositions (e.g., ``left of'' vs ``behind''). Instead, they manipulate the underlying semantic logic directly. This shifts the interaction paradigm from a tedious, linguistic trial-and-error loop into an interpretable, visual workflow.

However, the system relies on three image generation models, Gemini 2.5 Flash, Qwen Image Editing, and FLUX 1 Kontext, inherently inheriting the limitations of these underlying backbones. At the current stage, SceneCraft does not automatically evaluate or distinguish the strengths and weaknesses of each model's output; the final image selection still requires manual review by the user. Furthermore, the system's success depends on the accuracy of Detic and Grounding DINO; if an object is missed during the initial automated parsing, the user cannot easily interact with it.

\section{Conclusion}
In this paper, we have developed SceneCraft, an interactive system capable of generating and editing images via editable scene graphs. Each user interaction with the visual graph is automatically translated into a precise prompt that guides the corresponding image editing models. By abstracting the complexities of prompt engineering into structured relational manipulations, SceneCraft provides a highly intuitive, user-centered control mechanism. Our evaluations confirm that this paradigm not only improves objective image quality and relational alignment but significantly enhances user agency, transparency, and subjective satisfaction during the creative process.

For future work, we plan to design an intelligent agent that can automatically determine which image generation model to use, based on the identified strengths and weaknesses of each model. This will enable a higher level of automation and improve the overall efficiency of the framework.




\bibliographystyle{ACM-Reference-Format}
\balance
\bibliography{ref}










\end{document}